\documentclass[conference]{IEEEtran}
\usepackage{times}

\usepackage[numbers]{natbib}
\usepackage{multicol}
\usepackage[bookmarks=true]{hyperref}
\usepackage{times}
\usepackage{authblk}
\usepackage{multicol}
\usepackage[bookmarks=true]{hyperref}
\usepackage{flushend} % balance two columns at the end
\usepackage{graphics} % for pdf, bitmapped graphics files\
\usepackage{lipsum}% http://ctan.org/pkg/lipsum
\usepackage{epsfig} % for postscript graphics files
\usepackage{mathptmx} % assumes new font selection scheme installed
\usepackage{amsmath} % assumes amsmath package installed
\usepackage{amssymb}  % assumes amsmath package installed
\usepackage{mathtools}
\usepackage{color,soul}
\usepackage{paralist}
\usepackage{tikz} % latex-drawn figures
\usepackage{pgfplots}
\usepackage{float} % figure positioning
\usepackage{subcaption}
\usepackage{tabularx,booktabs}
\usepackage{pdfpages}

\newcolumntype{Z}[0]{>{\hsize=2\hsize}X}%
\newcolumntype{s}[0]{>{\hsize=1.2\hsize}X}%
\newcolumntype{n}[0]{>{\centering\arraybackslash\hsize=1.8\hsize}X}%
\usepackage[colorinlistoftodos,prependcaption,textsize=smallsize]{todonotes}
\newcommand{\inv}{^{-1}}
\newcommand{\tr}{^{\!\top}}

\newcommand{\argmin}{\operatornamewithlimits{argmin}}
\newcommand{\diag}{\mathit{diag}}

\newcommand{\se}{\mathfrak{se}(3)}
\newcommand{\SE}{SE(3)}

\newcommand{\R} {{\rm I\!R}}

\usepackage{xcolor}

\newcommand{\bl}{{\bar l}}

\begin{document}
	
	% paper title
	%\title{DOT-SLAM: Pose Change Representation for Simultaneous \\Localization, Mapping and Tracking in \\Dynamic Environments}
	%\title{Pose Change Representation for Dynamic Objects Tracking and Simultaneous Localization and Mapping }
	%\title{Pose Change Representation for Simultaneous Localization and Mapping with Dynamic Objects}
	\title{Simultaneous Localization and Mapping with Dynamic Rigid Objects}
	
	% You will get a Paper-ID when submitting a pdf file to the conference system
	\author{Mina Henein, Gerard Kennedy, Viorela Ila and Robert Mahony
	%\thanks{The authors are with the Australian Centre of Excellence for Robotic Vision, Research School of Engineering, The Australian National University, Canberra ACT 2601, Australia
	%{\tt\small mina.henein, gerard.kennedy, viorela.ila, and robert.mahony@anu.edu.au}}
	}
	\affil{The Australian Centre of Excellence for Robotic Vision,\\ Research School of Engineering,\\ The Australian National University,\\ Canberra ACT 2601,\\ Australia.\\
	mina.henein, gerard.kennedy, viorela.ila, and robert.mahony@anu.edu.au}

\maketitle

\begin{abstract}
Accurate estimation of the environment structure simultaneously with the robot pose is a key capability of autonomous robotic vehicles. Classical simultaneous localization and mapping (SLAM) algorithms rely on the static world assumption to formulate the estimation problem, however, the real world has a significant amount of dynamics that can be exploited for a more accurate localization and versatile representation of the environment. %to enable autonomous execution of tasks.
In this paper we propose a technique to integrate the motion of dynamic objects into the SLAM estimation problem, without the necessity of estimating the pose or the geometry of the objects. To this end, we introduce a novel representation of the pose change of rigid bodies in motion and show the benefits of integrating such information when performing SLAM in dynamic environments. Our experiments show consistent improvement in robot localization and mapping accuracy when using a simple constant motion assumption, even for objects whose motion slightly violates this assumption.

% The proposed system benefits robot localization in cluttered en

%objects in the formulation of the Simultaneous Localization, Mapping and Tracking in Dynamic Environments. In our experiments we show that integrating prior information about the dynamics of a rigid body into the estimation problem can increase the accuracy of the resulting map and robot localization.
	%The approach  is evaluated on a simulated environment, and a real indoor scenario. %We highlight the advantages of the proposed solution over SLAM formulation considering no dynamics in the environment.
\end{abstract}

\IEEEpeerreviewmaketitle

\section{Introduction}
% Motivation --> Claim --> Road Map

% Motivation
%In order to perform autonomous tasks such as surveying and inspection, search and rescue within man-made environments, de-mining, underwater monitoring, terrain mapping for localization in space and autonomous driving in urban environments, a robot must be able to simultaneously build an accurate representation of the surroundings and localise itself.

Autonomous robotic platforms are rapidly departing from simple, controlled environments to more complex and dynamic ones. Self driving cars, mobile or flying robots in urban environments are only a few examples of emerging applications of robotic in dynamic environments. In order to perform tasks autonomously, a robot must be able to simultaneously build an accurate representation of the surroundings and localize itself. Simultaneous localization and mapping (SLAM) algorithms are a core enabling technology for autonomous mobile robotics. %They have to be run on-board the robotic platform and provide a reliable solution at every step.
%These robotic platforms should also be able to work and interact in dynamic and cluttered environments.
%%Simultaneous localization and mapping is a mature field in robotics. Efficient algorithms exist which are able to accurately localize a robot and build a map of the environment. However, the existing techniques heavily rely on the assumption that most part of the environment is static. The moving objects are normally either removed from the estimation problem or track them separately. The problem of SLAM has been, in general, decoupled from the object tracking.

% Dynamic in SLAM
 SLAM is a well researched area in robotics, and many efficient solutions to the problem exist. However, the existing techniques heavily rely on the assumption that most parts of the environment are static~\cite{walcott12icra}.
 % , and many efficient solutions to the problem exist, however most of the them rely on the static world assumption~\cite{walcott12icra}. 
Nevertheless, real world scenarios have a significant amount of dynamics that should not be discarded. The conventional technique for dealing with dynamic environment in SLAM is to either treat any feature associated with moving objects as outliers~\cite{Hahnel02iros,Hahnel03icra,Wolf04icra,Wolf05autonrobot,Zhao08icra} or detect moving objects and track them separately using traditional multi-target tracking~\cite{Wang03icra,Miller07icra,Montesano08autonrobot,Rogers10iros}. However, ignoring dynamics in applications where the environment is cluttered with moving objects can cause the whole SLAM process to fail.

A typical SLAM system is composed of a front-end module, which processes the raw data from the sensors and a back-end module, which integrates the obtained information into a probabilistic framework. In visual SLAM, where cameras are the primary sensors for localization and mapping, simple primitives, such as point landmarks are commonly used to represent the environment. Points are easy to detect, track and integrate within the SLAM problem. Most of the existing methods assume that the dominant part of the scene changes only due to the camera motion, therefore they are prone to failure in highly dynamic environments. Other primitives such as lines and planes ~\cite{De14autonrobot,Kaess15icra,Henein17iros,Hsiao17icra} and even objects ~\cite{Mu16iros,Salas13cvpr} can provide richer map representations but they require higher level data processing. 

%In visual SLAM, where cameras are the primary sensors for localization and mapping, features (points, lines) detected from the images are associated and tracked over time. If such features are detected on a moving object,
%Moving objects can cause wrong data association, failure in landmark detection, failure in loop closing events as a result of occlusions and loss of observability of 3D points, and consequently, divergence in the state estimation ~\cite{zamora13tr}.\TODO{needs to be changed}.

% Sensing
Rapid advances in learning techniques for processing information from visual data, enable higher level sensing capabilities in robotic applications. %Objects can be easily identified and tracked.
Considerable effort has been devoted to developing accurate multi-object segmentation and tracking. Most of the existing techniques perform
tracking only in the image-domain~\cite{Xiang15iccv,Zhang08cvpr}, where pixels are labeled according to the object class they pertain to.
In most of the cases, those methods require additional processing in order to be integrated into robotic applications ~\cite{Salas13cvpr}. Other techniques use depth information obtained either from stereo cameras or LiDAR devices or scene geometry to track objects in 3D~\cite{Leibe07cvpr,Hoiem08ijcv}. In general, those methods require expensive point cloud processing, a priori models of the objects or even the known sensors locations.

In this paper, we propose a SLAM technique that can easily integrate information about the moving objects in the scene without the necessity of actually estimating the pose of the objects or relying on their 3D model. In particular, the formulation we propose uses the prior information that some of the structure points belong to moving objects, and the fact that these points can be associated in subsequent frames. In other words, it relies on a front-end able to identify and segment objects at image-level, associate points across the sequence of images and use semantic information to classify the feature points into either static or dynamic. It is worth mentioning that the problems of instance level object segmentation and data association are out of the scope of this paper, and considered as future work. The main contribution of this paper is the use of a novel \emph{pose change representation} to model the motion of the points pertaining to rigid bodies in motion and the integration of such model into a SLAM optimization framework. This has the important benefits of keeping the formulation simple and of using minimal information about the objects. Furthermore, the formulation also has the flexibility to accommodate prior information about the motion of the objects. In particular we analyze the effect of using a \emph{constant motion} model assumption to model the dynamic objects in the environment. We use the term ``constant motion'' to indicate that pose transformation is conserved. The constant motion assumption can be used for objects that appear in the field of view for a short period of time. In this paper we will demonstrate the effects of this assumption in a tracking and mapping estimation problem.

The remainder of this paper is structured as follows, in the following section we discuss the related work. In section \ref{sec:Problem} we describe the proposed approach for incorporating dynamics of the scene. In section \ref{sec:Setup} we introduce the experimental setup, followed by the experimental results and evaluations in section \ref{sec:Experiments}. We summarize and offer concluding remarks in section \ref{sec:Conclusion}.

%%%%%
%In this paper, we explore the effect of integrating additional information about the points in the environment into the SLAM estimation problem. The type of information we consider is the fact that some of the observed points pertain to moving objects.  %information that a number of landmarks belong to the same rigid object and the object can . without the need of explicitly introducing the pose of the object as a random variable,

%\textbf{[RM : This material is better suited to the introduction]}
%In this paper we formulate a simultaneous robot localization, environment mapping and tracking of moving objects problem with applications to robot navigation in dynamic environments.

%For simplification, in this paper we assume that the objects are moving with constant motions and the point association to moving objects and static scene is given.

%%%%%%%%%%%%%%%%%%%%%%%%% Related Work
\section{Related Work}
\label{sec:Litreview}

% KF SLAM
The earliest work on SLAM was based on the extended Kalman filter (EKF) approach~\cite{Cheeseman87isrr} \cite{Leonard92ijrr} \cite{Ayache89tra}.
However, it has been shown that filtering is inconsistent when applied to the inherently non-linear SLAM problem~\cite{Julier01icra}.
% Graph-based SLAM
One intuitive way of formulating SLAM is to use a graph representation with nodes corresponding to the random variables (robot poses and/or landmarks in the environment) and edges representing functions of those variables (typically measurement functions). Lu and Milios ~\cite{Lu97autonrobot} first proposed the graph-based formulation of the SLAM problem in 1997.
Once such a graph is constructed, the goal is to find a configuration of the nodes that is maximally consistent with the measurements~\cite{Grisetti10itsm}.
Approaching SLAM as a non-linear optimisation on graphs has been shown to offer very efficient solutions to SLAM applications ~\cite{Grisetti07rss,Kummerle11icra,Ila15icra}.
%Factor graphs, such as the one in Figure\ref{fig:SLAMGraph}, are graphical models that have been used for representing the SLAM problem~\cite{Dellaert06ijrr} \cite{Kaess08tro} \cite{Kaess11ijrr}. This is due to the fact that in factor graphs the functions are made explicit and such a bipartite graph is directly connected to the solutions of the optimisation problem~\cite{Kaess11ijrr}.
%\input{figures/graph}

%Accounting for dynamics in SLAM
Establishing the spatial and temporal relationships between a robot, stationary objects, and moving objects in a scene serves as a basis for scene understanding~\cite{Wang07ijrr}. Simultaneous localization, mapping and moving object tracking are mutually beneficial. However, in SLAM algorithms, information associated with stationary objects is considered positive, while moving objects are seen to degrade the performance. Conversely, measurements belonging to moving objects are required for moving object tracking algorithms, while stationary points and objects are considered background and filtered out.

The conventional technique for dealing with dynamic objects in SLAM is to detect them and then either treat them as outliers~\cite{Hahnel02iros,Hahnel03icra,Wolf04icra,Wolf05autonrobot,Zhao08icra} or track them separately using traditional multi-target tracking ~\cite{Wang03icra,Miller07icra,Montesano08autonrobot,Rogers10iros}.
Hahnel \emph{et al.} ~\cite{Hahnel03icra} used an Expectation-Maximization (EM) algorithm to update the probabilistic estimate about which measurements corresponded to a static/dynamic object and removed them from the estimation when they corresponded to a dynamic object. Bibby and Reid's SLAMIDE ~\cite{bibby07rss} also estimates the state of 3D features (stationary or dynamic) with a generalised EM algorithm where they use reversible data association to include dynamic objects into a single framework SLAM.  %Bleser et al. ~\cite{bleser06ismar} propose a camera tracking system similar to monoSLAM ~\cite{davison07tpami}, which will delete features from the map if they have not been tracked in more than 50 percent of the frames where they should be visible (as predicted by camera poses). Under the framework of SfMbased SLAM, such as PTAM, Shimamura et al.'s vSLAM ~\cite{shimamura11mva} estimates the flow vector throughout all the outliers after camera pose estimation on the tracking thread, and clusters the flow vectors by Gaussian mixture models (GMM). If the number of outliers in a cluster exceeds a certain threshold, the outliers are considered to be on moving objects and are eliminated from the map.

Wang \emph{et al.} ~\cite{Wang07ijrr} developed a theory for performing SLAM with Moving Objects Tracking (SLAMMOT). They first presented a SLAM algorithm with generalised objects, which computes the joint posterior over all objects and the robot, an approach that is computationally demanding and generally infeasible as stated by the authors. They also developed a SLAM with detection and tracking of moving objects, in which the estimation problem is decomposed into two separate estimators, for the stationary and moving objects, which results in a much lower dimensionality and makes it feasible to update both filters in real time. In computer vision, there has been some progress in developing systems for multi-object tracking from moving cameras~\cite{Leibe07cvpr,Hoiem08ijcv}, but in general those methods focus on the object tracking problem and require visual odometry or known camera poses.

%MINA - removing as per Rob's comment
%Different to the above mentioned methods, our approach is designed to integrate and solve in a single framework the whole problem of simultaneous localization, mapping and moving objects tracking as they are mutually beneficial and utilising the information from each of them improves the solution of the other.

%%%%%%%%%%%%%%%%%%%%%%%%% Problem Formulation
\section{Accounting for dynamic objects in SLAM}
\label{sec:Problem}

%The problem considered is one in which there are \emph{relatively large} rigid-objects moving with \emph{constant motion} within the sensing domain of the robot that is undertaking the SLAM optimisation.
%By \emph{relatively large} we mean that the SLAM front-end system identifies many points from the same rigid-body object.
%That is, many points in the SLAM optimisation problem share an underlying motion constraint that can be exploited to improve the quality of the SLAM estimation.
%There can, of course, be several or many rigid-bodies moving independently, however, the benefit of the proposed approach is only seen when several point features from each object are observed.
%By \emph{constant motion} we mean that the body-fixed frame pose change (see \eqref{eq:XXXXX} below) is constant.
%That is, from the point of view of the object, it moves with a constant body-fixed frame velocity.  \TODO{Move this where the }
%
%We assume additionally that the SLAM front-end is able to provide a segmentation of the data, that is, it can associate each observed feature with a given object.
%We do not require that SLAM front-end has any further understanding of the rigid-body object.
%In particular, no pose estimate is made by the front-end, and indeed, no geometric model of the object is required by the algorithm.

The problem considered is one in which there are relatively large rigid-objects moving within the sensing domain of the robot that is undertaking the SLAM estimation.
The SLAM front-end system is able to identify and associate points from the same rigid-body object at different time steps.
These points share an underlying motion constraint that can be exploited to improve the quality of the SLAM estimation. 
The proposed approach does not require that the SLAM front-end has any understanding of the rigid-body object beyond the ability to segment and associate feature points associated with the objects. 
In particular, no pose estimate is made by the front-end, and indeed, no geometric model of the object is required by the algorithm. 
The following sections will show how the motion of the rigid body object can be directly transferred to the motion of the feature points and how this information can be integrated into the SLAM estimation.

%~~~~~~~~~~~~~~~~~~~~~~~~~~~~~~~~~~~~~~~~~~~~~~~~~~~~~~~~~~~~~~~~~~~~~~~~~~~~~~
\subsection{Motion model of a point on a rigid body}
\begin{figure}[t]
\centering
\includegraphics[width=.85\linewidth,trim=0mm 145mm 0mm 0mm,clip]{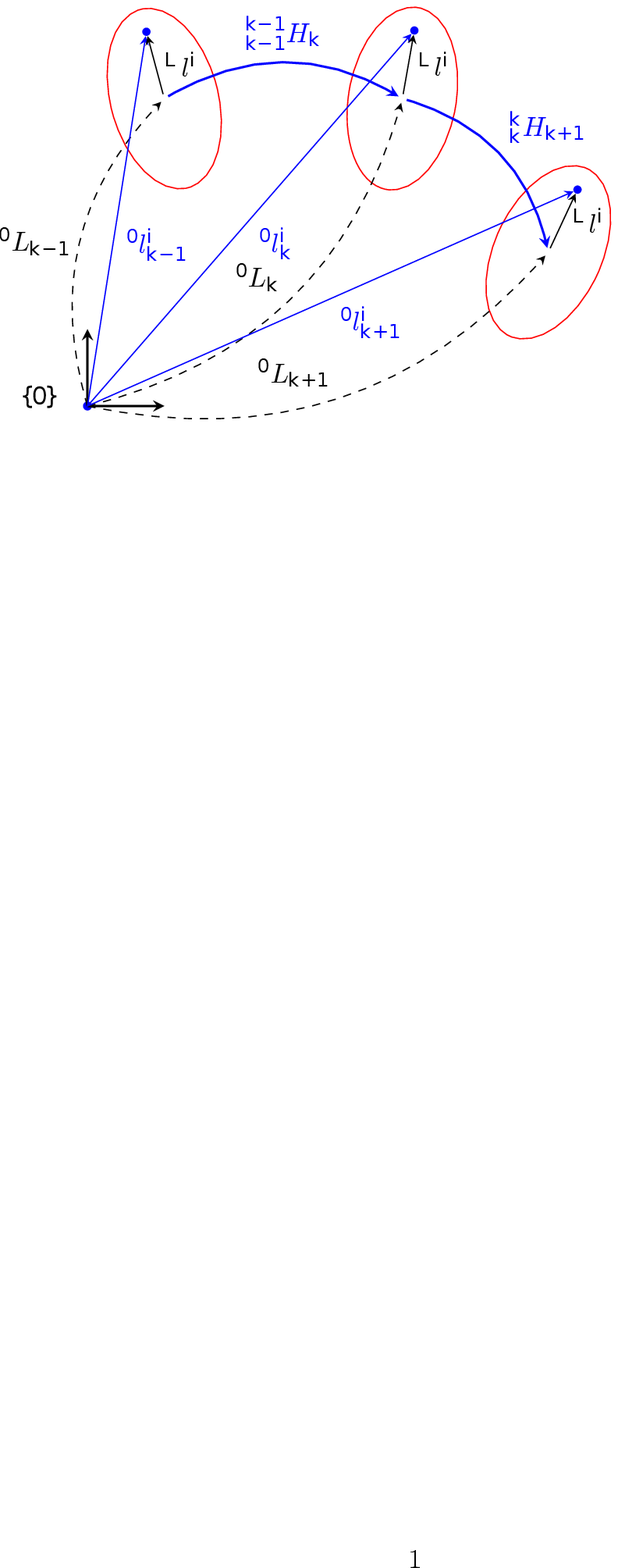}
\caption{Representative coordinates of the rigid body in motion. The points $^Ll^i$ are represented relative to the rigid body centre of mass $\{L\}$ at each step. }
\label{fig:CoordinatesRigidBody}
\end{figure}
Let $\{0\}$ denote the \emph{reference coordinate frame}, and $\{L\}$ a coordinate frame associated to a moving rigid body (object).
We write the pose $^0 L_k\in \SE$ of the rigid-body as an element of $\SE$ with respect to the reference frame $\{0\}$.
For feature observed on an object, let ${^L} l^i \in \R^3$ denote the coordinates of this point in the body-fixed frame, where `$i$' is the unique index of the feature.
We write ${^0}l^i_k$ for the coordinates of the same point expressed in the reference frame $\{0\}$ at time $k$.
Note that for rigid bodies in motion, ${^L} l^i$ is constant for all the object instances, while both $^0 L_k$ and ${^0}l^i_k$ are time varying.
The point coordinates are related by the expression:
\begin{eqnarray}
^L\bl^i =\:  ^0 L_k\inv \: ^0\bl_k^i \label{eq:TransPoint}\:.
 \label{eq:RelPoint}
\end{eqnarray}
where ${^L}\bl^i$ and ${^L}\bl^i_k$ are the homogeneous coordinates of the points ${^L} l^i$ and ${^0}l^i_k$, respectively.

%In this work we assume rigid bodies in motion, which means that the point relative to the object frame remains the same at each time step:
%%
%\begin{eqnarray}
%^0_0 L_{k}\inv\:  ^0\bl_{k}^i=\:^L\bl_k^i = \:^L\bl_{k+1}^i = \:^0 L_{k+1}\inv\:  ^0\bl_{k+1}^i \label{eq:RigidBody}\:.
%\end{eqnarray}
%%

The relative motion of the object $L$ from time $k$ to time $k+1$ is represented by a rigid-body transformation $_k^kH_{k+1}\in \SE$ that we term the \emph{body-fixed frame  pose change}.
Here the lower indices indicate that the transformation maps a base pose  (lower left index) to a target pose lower right index, expressed in coordinates of the base frame (upper left index).  
The body-fixed frame pose change is the $\SE$ transformation:
\begin{eqnarray}
_k^k H_{k+1} = ^0 L_k\inv {^0 L_{k+1}}  
 \label{eq:BFFIC}
\end{eqnarray}
that is `classical' in most robotics developments. Figure \ref{fig:CoordinatesRigidBody} shows this transformations for three consecutive object poses. In consequence, relative motion of the rigid body is given by the incremental pose transformation:
\begin{eqnarray}
{^0 L_{k+1}}  = {^0 L_k}\, {^k_k H_{k+1}}\:.
 \label{eq:RelMotion}
\end{eqnarray}
Consider a point $^L l^i$ in the object frame $\{L\}$. The motion of this point can be obtained by writing the expression \eqref{eq:RelPoint} for two consecutive poses of the object at time $k$ and $k+1$ and using the relative motion of the object in \eqref{eq:RelMotion}. With that we obtain:
\begin{eqnarray}
^0\bl_{k+1}^i =\: ^0 L_{k}  \:^k_k H_{k+1}  \:^0 L_{k}\inv \:  ^0\bl_{k}^i \:.
\label{eq:PointMotion}
\end{eqnarray}
We observe that \eqref{eq:PointMotion} relates points on the same rigid body in motion at different time step by a transformation \mbox{$^0_kH_{k+1} = \:^0 L_{k} \:  ^k_kH_{k+1}\: ^0 L_{k}\inv $}, where \mbox{$^0_kH_{k+1}\in \SE$}. According to ~\cite{Chirikjian17idetc}, this equation represents a frame change of a pose transformation, and shows how the body-fixed frame pose change in \eqref{eq:BFFIC} relates to the \emph{reference frame pose change}. The point motion model in the reference frame becomes:
\begin{eqnarray}
 ^0\bl_{k+1}^i =\: ^0_kH_{k+1} \: ^0\bl_{k}^i \:.
 \label{eq:ifTransform}
\end{eqnarray}
This formulation is key to the proposed approach since it eliminates the need to estimate the actual object pose $^0 L_{k}$ and allows us to work directly with points ${^0 \bl^i_k}$ in the reference frame.

Assuming the objects are moving with \emph{constant velocity} is a reasonable first approximation of the motion of many moving objects such as cars, trucks, bicycles, etc, that are of interest in real world SLAM problems. The constant motion assumption is natural to pose in the body-fixed pose change: 
\begin{eqnarray}
{^k_k H_{k+1}} = C = {^{k'}_{k'} H_{{k'}+1}} \in \SE
\label{eq:ConstantC}
\end{eqnarray}
for any $k, k'$ time indices. 

A second key observation we make in this paper is that if the body-fixed frame pose change is constant then the reference frame pose change is constant too.
To see this, we rescale \eqref{eq:RelMotion} and use \eqref{eq:ConstantC} to obtain:
\begin{eqnarray}
^0 L_{k}  = ^0 L_{k-1}\, C
\end{eqnarray}
which we replace in $^0_kH_{k+1} = \:^0 L_{k} \:  C\: ^0 L_{k}\inv$ to obtain:
\begin{eqnarray}
^0_kH_{k+1} = ^0 L_{k-1} \, C  \:^0 L_{k-1}\inv =\:^0_{k-1} H_{k}
\end{eqnarray}
%and \eqref{eq:ConstantC} to obtain:
%\begin{eqnarray}
%{^0 L_{k+1}} & = {^0 L_{k-1}} \, C^2 = {^0 L_{k-1}}\, C \,{^0 L_{k-1}\inv} \, {^0 L_{k}} 
%\end{eqnarray}
%%
%and the fact that \mbox{$^0_{k-1}H_{k} = \:^0 L_{k-1} \:  C\: ^0 L_{k-1}\inv $} to obtain:
%\begin{eqnarray}
% {^0 L_{k+1}}   = {^0_{k-1} H_{k}} \, {^0 L_{k}}. 
% \end{eqnarray}
% %
%Solving for ${^0_{k-1} H_{k}}$ and recalling \eqref{eq:RelMotion} one has 
%\begin{eqnarray}
%{^0_{k-1} H_{k}} = {^0 L_{k+1}} {^0 L_{k}\inv} = {^0_{k} H_{k+1}}. 
%\end{eqnarray}
%%
It follows that the reference frame pose change:
\begin{eqnarray}
{^0_{k-1} H_{k}} = H = {^0_{k'} H_{{k'}+1}} \in \SE
\end{eqnarray}
holds for any $k, k'$ indices. Therefore, for a rigid-body object in motion we can use a constant reference frame pose change $H \in \SE$ that acts on the points on the rigid body to update their reference frame coordinates: $^0\bl_{k+1}^i = H{^0 \bl^i_k}$.

\begin{figure}[t]
\centering
\includegraphics[width=.82\linewidth,trim=0mm 125mm 0mm 0mm,clip]{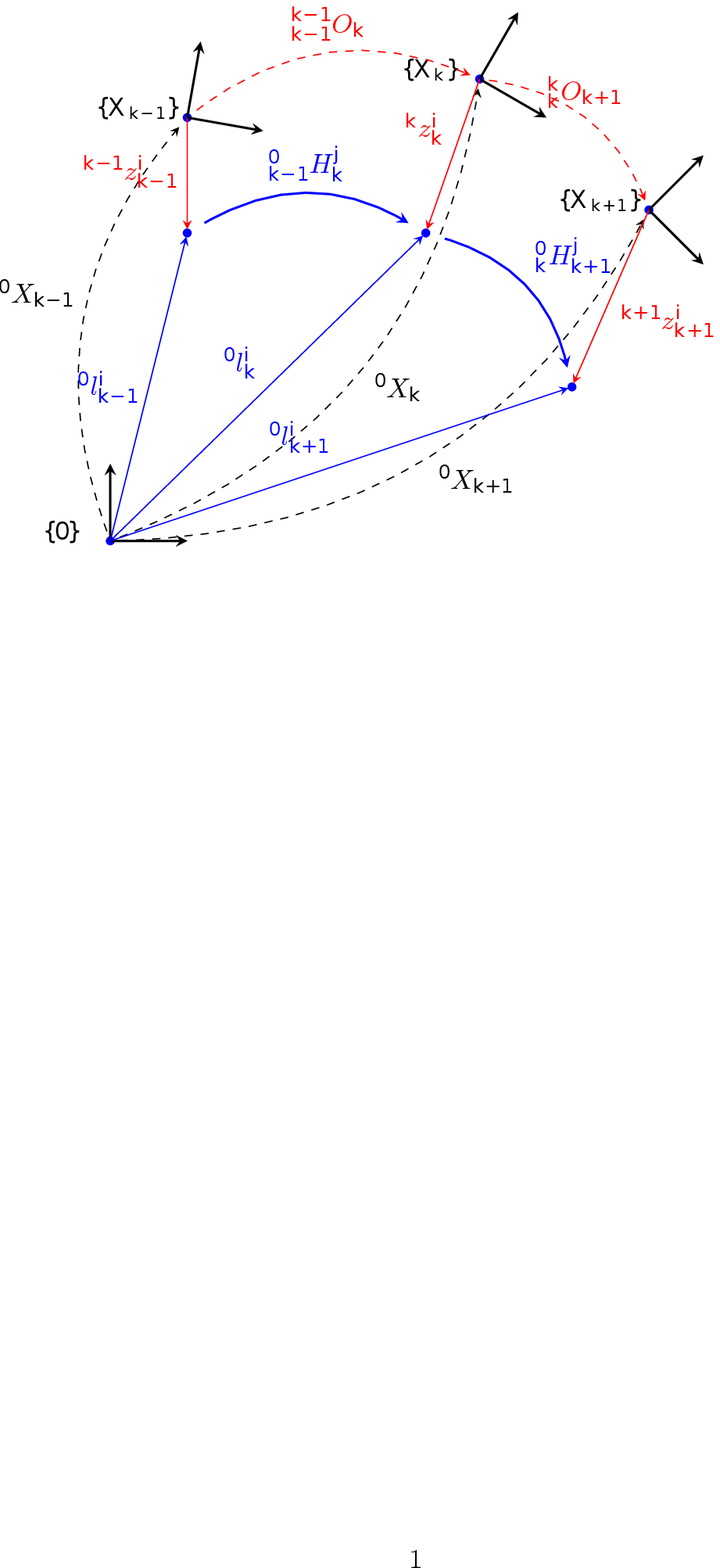}
\caption{Representative coordinates of a system with a rigid body in motion. Blue represents state elements and red the measurements. For consistency with Figure \ref{fig:CoordinatesRigidBody}, in here we show all the coordinates and relative motions in $\SE$, however, in the actual state vector we use the corresponding logarithm map of $\SE$.}
\label{fig:Coordinates}
\end{figure}
\subsection{State representation}

The SLAM with dynamic objects estimation problem is modelled using factor graphs~\cite{Dellaert06ijrr}, and the goal is to obtain the 3D structure of the environment (static and dynamic objects) and the robot poses that maximally satisfy a set of measurements and motion constraints. Assuming Gaussian noise, this problem becomes a non-linear least squares (NLS) optimisation over a set of variables~\cite{Polok13icra}: the camera/robot poses \mbox{$\textbf{x} = \{x_0 ... x_{n_x}\}$}, with $x_k\in \se$, the tangent space of $\SE$ at identity, where $k \in 0 ... n_x$ and $n_x$ is the number of steps,
% (with \mbox{ $\{x_k \in \se  |\: x_k= \log(X_k) \}$, $\log(\cdot)$} being the exponential map of the Special Euclidian group $\SE$);
 and the 3D point features in the environment seen at different time steps:  \mbox{${\bf l} = \{l^1_0 \dots \:l^{n_l}_{n_x}\}$} where \mbox{$l^i_k \in \R^3$ } and $i \in 1 ... n_l$ is the unique index of a landmark and $n_l$ is the total number of detected landmarks. 
 
 The set of landmarks, ${\bf l} = {\bf l_s} \cup {\bf l_d}$, contains a set of static landmarks ${\bf l_s}$ and a set of landmarks detected on moving objects at different time steps, ${\bf l_d}$. The formulation in this section assumes that the static/dynamic classification and the association of the points at different time steps is done by the front-end. The same point on a moving object is represented using a different variable at each time step, i.e. $l^i_{k-1}$ and $l^i_k$ are the same physical point seen at time $k-1$ and $k$, respectively. In this particular problem formulation, the robot/camera poses and the positions of the 3D points are represented in the $\{0\}$ reference frame, which is omitted from their notation in this section. Nevertheless, the application of the technique is general and can easily accomodate body-fixed frame robot poses and relative points\cite{Polok15acra}.
 
%. For the clarity of the presentation, in this section, we omit the reference frame from the notation the landmark points and robot poses.
%We also consider the camera and all the other sensors centred at the body-fixed frame. In practice, sensor calibration is required in order to establish the deviations from the body-fixed frame.

%A single variable is used to represent each static point and the front-end is able to track and associate those points.
Assuming rigid body objects moving in the environment, points on the same object have similar motion. Therefore, in the proposed formulation, we integrate a new type of state variable characterising this motion. Equation \eqref{eq:ifTransform} allows us to relate the points on moving rigid bodies at different time steps ($k$ and $k+1$) with a reference frame pose change: $^0_kH_{k+1}\in \SE$. %Likewise, we omit the reference frame $\{0\}$ from top-left corner of the notation, and we will further refer this pose change as $_kH_{k+1}$. 
To avoid over-parameterization, the logarithm map of an $\SE$ element is used as a state variable, \mbox{$\{^0_k u _{k+1}^j \in \se |^0_k u _{k+1}^j= \log(\:^0_kH_{k+1}^j)\}$}, where $j\in 1 ... n_o$ is the object index and $n_o$ is the number of identified objects. The set of all variables is now ${\bf \theta} = {\bf x} \cup {\bf l} \cup{\bf u}$, where ${\bf u}$ is the set of all the variables characterising the objects' motion.

%
%\begin{figure}[t]
%\centering
%\includegraphics[width=1.0\linewidth,trim=50mm 200mm 70mm 40mm,clip]{./figures/SE3MotionVertexGraph.pdf}
%\caption{Factor graph representation for SLAM problem with an SE(3) motion vertex and its edges. Superscripts represent the same landmark at different time steps.}\label{fig:ManySE3MotionVerticesGraph}
%\end{figure}
%%
\begin{figure}
\centering
	\includegraphics[width=.85\linewidth,trim=0mm 173mm 0mm 0mm,clip]{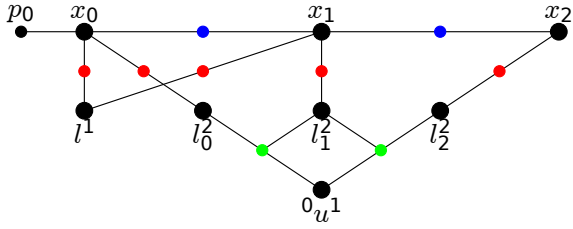}
\caption{Factor graph representation of a problem with an SE(3) motion vertex and its edges. Superscripts represent the same landmark at different time steps. Blue factors represent odometric measurements, red factors represent point measurements in camera frames and green factors represent `motion factors'}
\label{fig:SE3MotionGraphs}
\end{figure}

\subsection{Measurements and object motion constraints}

Two types of measurements, the odometry obtained from the robot's proprioceptive sensors and the observations of the landmarks in the environment obtained by processing the images from an on-board camera are typically integrated into a visual SLAM application.
Let \mbox{$f(x_{k-1},x_{k})$} be the odometry model with $\Sigma_{v_k}$, odometry noise covariance matrix:
\begin{eqnarray}
 o_k = f(x_{k-1},x_{k})+ v_k\:, & \text{with} & v_k \sim \mathcal{N}(0,\Sigma_{v_k})\:,
\label{eq:odoModel}
\end{eqnarray}
and \mbox{$\textbf{o} = \{o_1 ... o_{m_i}\} $} being the set of ${m_i}$ odometric measurements.
Similarly, let \mbox{$h(x_k,l^i_k)$} be the 3D point measurement model with $\Sigma_{w_k}$, the measurement noise covariance matrix:
\begin{eqnarray}
z_k^i = h(x_k,l^i_k)+ w_k^i\:, & \text{with} & w_k^i \sim \mathcal{N}(0,\Sigma_{w_k})
\label{eq:pointModel}
\end{eqnarray}
where \mbox{$\textbf{z} = \{z_1 ... z_{m_k}\}$, $z_k \in {\rm I\!R}^3$} is the set of all 3D point measurements at all time steps. Figure \ref{fig:Coordinates} shows the measurements in red. %Observe that we deliberately omitted the left superscripts of the landmarks and robot poses variables, as we would like to keep the formulation general, their state representation can be either in inertial or relative coordinates.

The relative pose transformation of the points on moving rigid bodies is given in \eqref{eq:ifTransform}. Observe that the motion of any point on a specific object $j$ can be characterised by the same pose transformation $^0_kH_{k+1}^j \in \SE$ with $^0_kR_{k+1}^j$ the rotation component and $^0_k t_{k+1}^j$ the translation component, respectively. The factor corresponding to \eqref{eq:ifTransform} is:
\begin{eqnarray}
g(l^i_k,l^i_{k+1},^0_k u _{k+1}^j) = { ^0_kR_{k+1}^j}\tr \: ^0l_{k+1}^i - {^0_kR_{k+1}^j} \tr\:{^0_k t_{k+1}^j} -^0l_k^i  + q_{s_j}
\label{eq:constantMotionGeneral}
\end{eqnarray}
with \mbox{$ ^0_kR_{k+1}^j $} and \mbox{${^0_k t_{k+1}^j}$} obtained using the exponential map $\exp(^0_k u _{k+1}^j)$ and \mbox{$q_s \sim \mathcal{N}(0,\Sigma_{q})$} is the normally distributed zero-mean Gaussian noise. The factor in \ref{eq:constantMotionGeneral} is a ternary factor which we call the \emph{motion model of a point on a rigid body}. 

In this paper we analyze the effect on the solution of the SLAM problem of the assumption that the objects are moving with constant motion. Figure \ref{fig:SE3MotionGraphs} depicts the factor graph of a small SLAM example of three robot poses, a static feature and a feature detected at three different time steps on an object with constant motion. We say that a pose change $^0_k H _{k+1}^j$ is constant for all the points on an object $j$ at every time step, hence a sole state variable $^0 u ^j$ is used for each object and the factor in \eqref{eq:constantMotionGeneral} becomes:
\begin{eqnarray}
g(l^i_k,l^i_{k+1},^0 u^j) = { ^0R^j}\tr \: ^0l_{k+1}^i - {^0R^j} \tr\:{^0 t^j} -^0l_k^i  + q_{s_j}
\label{eq:constantMotionGeneral1}
\end{eqnarray}

%
%A relative motion $\exp(u)$ between the same landmarks seen at different time steps is characterized by a pose $\in \se$.
%For each point seen at two consecutive time instants $l^i$ that belongs to a rigid object with a relative motion $\exp(u)$ we have:
%\begin{eqnarray}
%0 = g(l^i_k,l^i_{k+1},\exp(u)) + q_{s_j} &\text{and}& \\
%g(l^i_k,l^i_{k+1},\exp(u)) = \prescript{0}{}{l^i_k} -  \prescript{0}{}{\exp(u)^{-1}} \prescript{0}{}{l^i_{k+1}}
%\end{eqnarray}

%and its jacobians are: \\
%\begin{equation}
%\frac{\partial g(l^i_k,l^i_{k+1},\exp(u))}{\partial l^i_k} = eye(4)
%\end{equation}
%\begin{equation}
%\frac{\partial g(l^i_k,l^i_{k+1},\exp(u))}{\partial l^i_{k+1}} = -  \prescript{0}{}{\exp(u)^{-1}}
%\end{equation}
%\begin{equation}
%\frac{\partial g(l^i_k,l^i_{k+1},\exp(u))}{\partial \exp(u)} = -  \prescript{0}{}{\exp(u)^{-1}} G_j \prescript{0}{}{l^i_{k+1}}
%\end{equation}

%\subsection{The graph optimization }

%\begin{figure}[h]
%		\centering
%		\includegraphics[height=1.4in,trim=4cm 8cm 4cm 8cm ,clip]{./figures/ExpA_Env}
%		\caption{\small \TODO{CHANGE THIS FIGURE: MINA -- for spy(A) of exp.A w/ and w/o lc}}
%		\label{fig:ExpBEnv}
%\end{figure}

\subsection{The graph optimization }
Given the measurements and motion factors introduced above, we can formulate an NLS problem to obtain the optimal solution of the SLAM with dynamic objects:
\begin{eqnarray}\nonumber
\boldsymbol \theta^* = \argmin_{\boldsymbol \theta}\: \frac{1}{2} \Bigg\{
\sum_{i=1}^{mi} \| f(x_{k-1},x_k) - o_k  \|_{\Sigma_{v_k}}^2  + \\
\sum_{k=1}^{mk} \| h(x_{i},l^i) - z_k\|_{\Sigma_{w_k}}^2 +
\sum_{i,j}^{ms} \| g(l^i_k,l^i_{k+1},^0u^j)  \|_{\Sigma_{q}}^2 \:\Bigg\}
\label{eq:NLSMC}
\end{eqnarray}
where $mi$, $mk$ and $ms$ are the number of odometric measurements, point measurement and constant motion factors.

Iterative methods such as Gauss-Newton (GN) or Levenberg-Marquard (LM) are used to find the solution of the NLS  in \eqref{eq:NLSMC}.
An iterative solver starts with an initial point $\boldsymbol \theta^0$ and, at each step, computes a correction $\boldsymbol \delta$ towards the solution.  For small $\left\Vert\boldsymbol \delta \right\Vert$, a Taylor series expansion leads to linear approximations in the neighborhood of $\boldsymbol \theta^0$ and a linear system \mbox{$A\tr A \delta =- A\tr \bf b$} is solved at each iteration ~\cite{Kaess08tro,Polok13icra}.
%\begin{equation}
%\delta^* = \argmin_{\delta}\: \frac{1}{2} \|{A\tr A \delta - A\tr \bf b}\|_{\Sigma}^2\:,
%\label{eq:LS}
%\end{equation}
In here, $A$ gathers the derivatives of all the factors in \eqref{eq:NLSMC} with respect to variables in $\boldsymbol \theta$ weighted by the square rooted covariances of each factor, and $b$ is the residual evaluated at the current linearization point. The new linearization point $\theta^{k+1} $ is obtained by applying the increment $\delta^*$ to the current liniarization point  \mbox{$ \boldsymbol \theta^i $}. This formulation is often used in the SLAM literature~\cite{Dellaert06ijrr,Kaess08tro,Kummerle11icra,Polok13icra}.

The factor graph formulation of the SLAM problem is highly intuitive and has the advantage that it allows for efficient implementations of batch~\cite{Dellaert06ijrr} \cite{ceres-solver} and incremental~\cite{Kaess11ijrr,Polok13rss,Ila17ijrr} NLS solvers. However the efficiency is highly reliant on the sparsity of the resulting graph. Introducing the constant motion vertex may affect the sparsity of the graph by connecting many landmark nodes to one single vertex. This can lead to inefficiency if it is not handled properly when solving the linear system. The importance of variable ordering when solving a linear system using matrix factorization has been studied in the SLAM literature \cite{Agarwal12iros,Polok13rss}. An appropriate permutation of the original matrix can be used, yielding small fill-in in the resulting factorization. This has significant computational advantages, since the back-substitution using a sparse triangular matrix is very efficient. Therefore, when solving the problem in \eqref{eq:NLSMC} we insure that the motion related variables are ordered last.

\begin{figure*}[h]
	\centering
		\begin{subfigure}[t]{0.19\textwidth}
		\centering
		\includegraphics[height=1.4in,trim=5cm 9cm 5cm 9cm  ,clip]{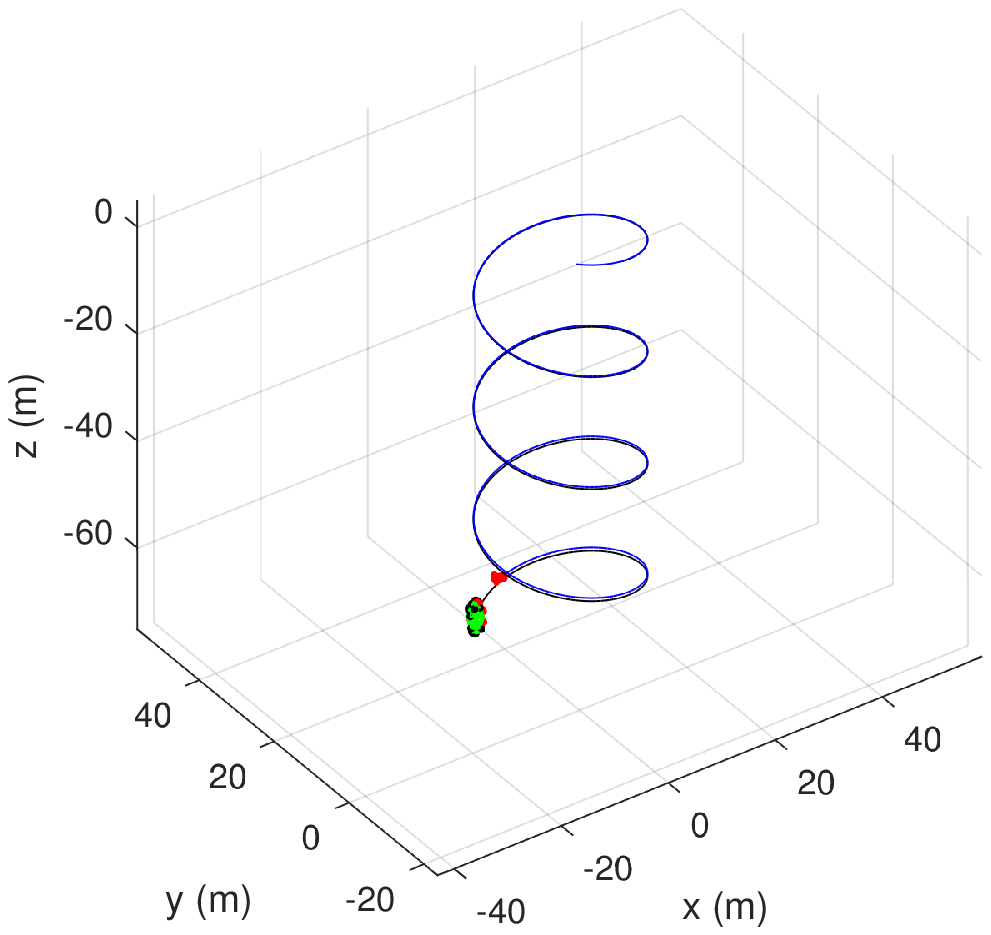}
		\caption{\small Experiment `A'}
		\label{fig:ExpAEnv}
	\end{subfigure}%
	\begin{subfigure}[t]{0.19\textwidth}
		\centering
		\includegraphics[height=1.4in,trim=5cm 9cm 5cm 9cm  ,clip]{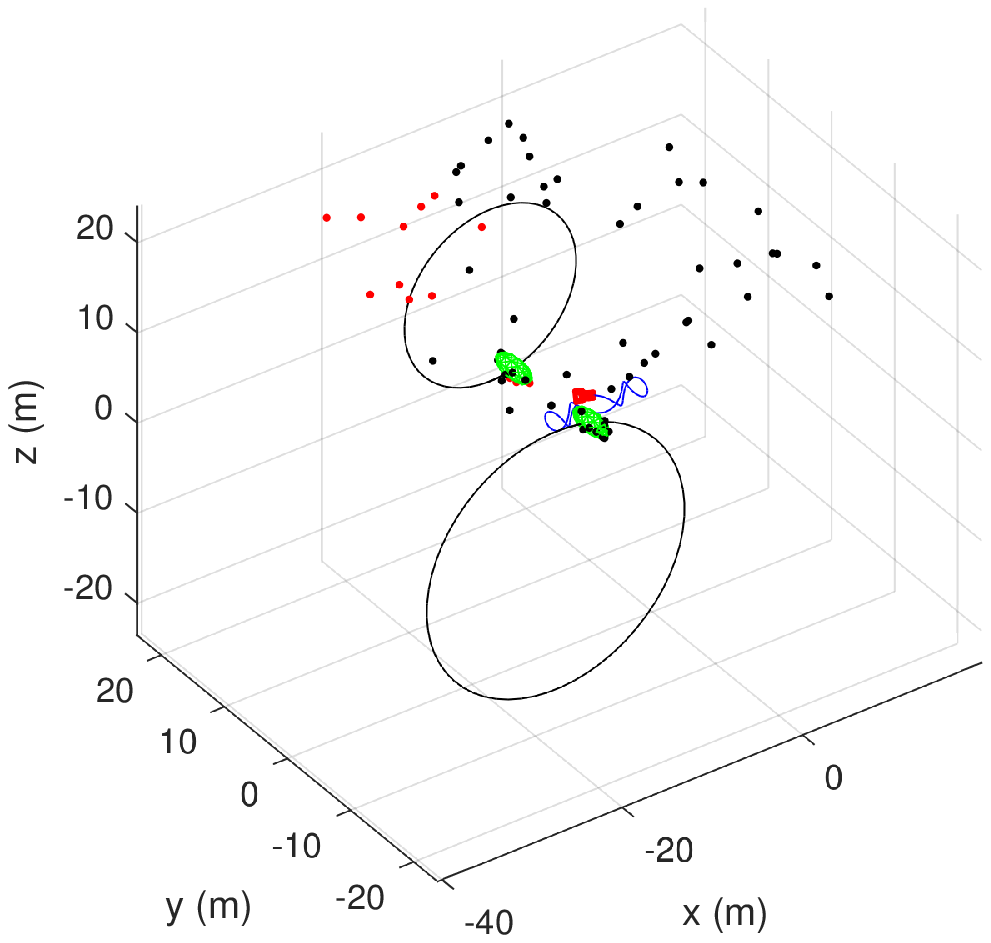}
		\caption{\small Experiment `B'}
		\label{fig:ExpBEnv}
	\end{subfigure}%
	\begin{subfigure}[t]{0.19\textwidth}
		\centering
		\includegraphics[height=1.4in,trim=5cm 9cm 5cm 9cm  ,clip]{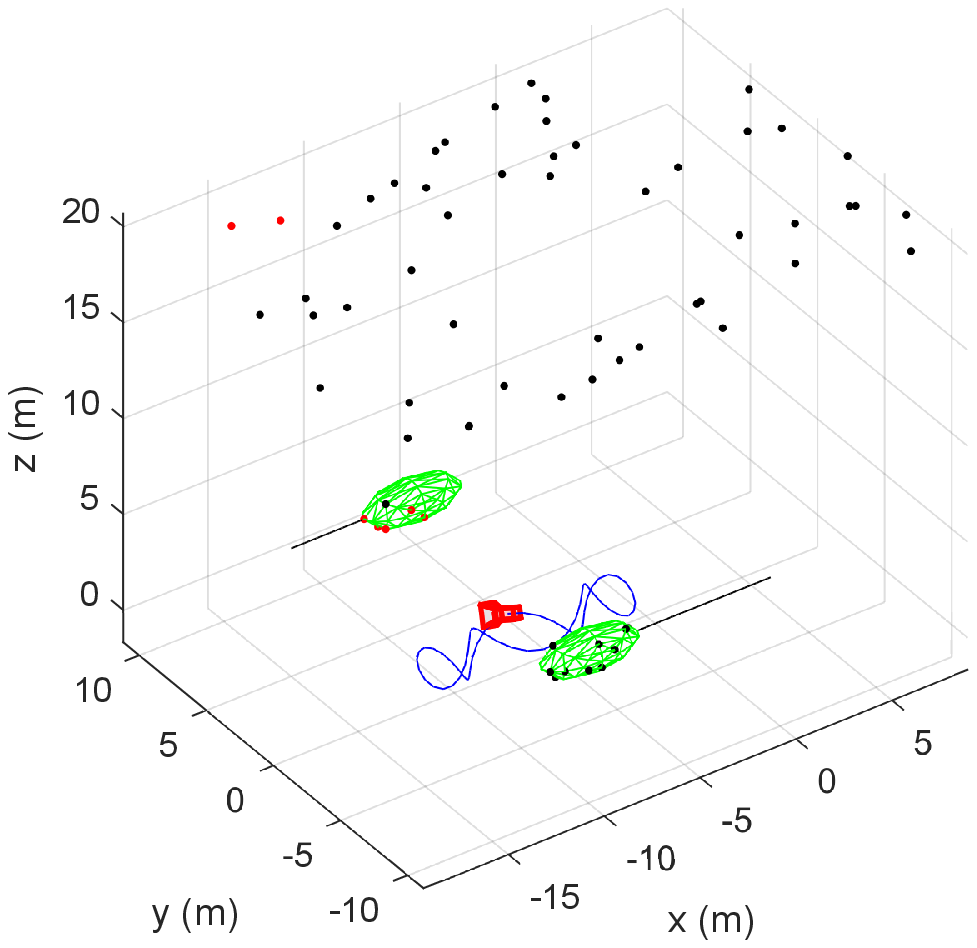}
		\caption{\small Experiment `C' }
		\label{fig:ExpCEnv}
	\end{subfigure}
	\begin{subfigure}[t]{0.19\textwidth}
		\centering
		\includegraphics[height=1.4in,trim=5cm 9cm 5cm 9cm ,clip]{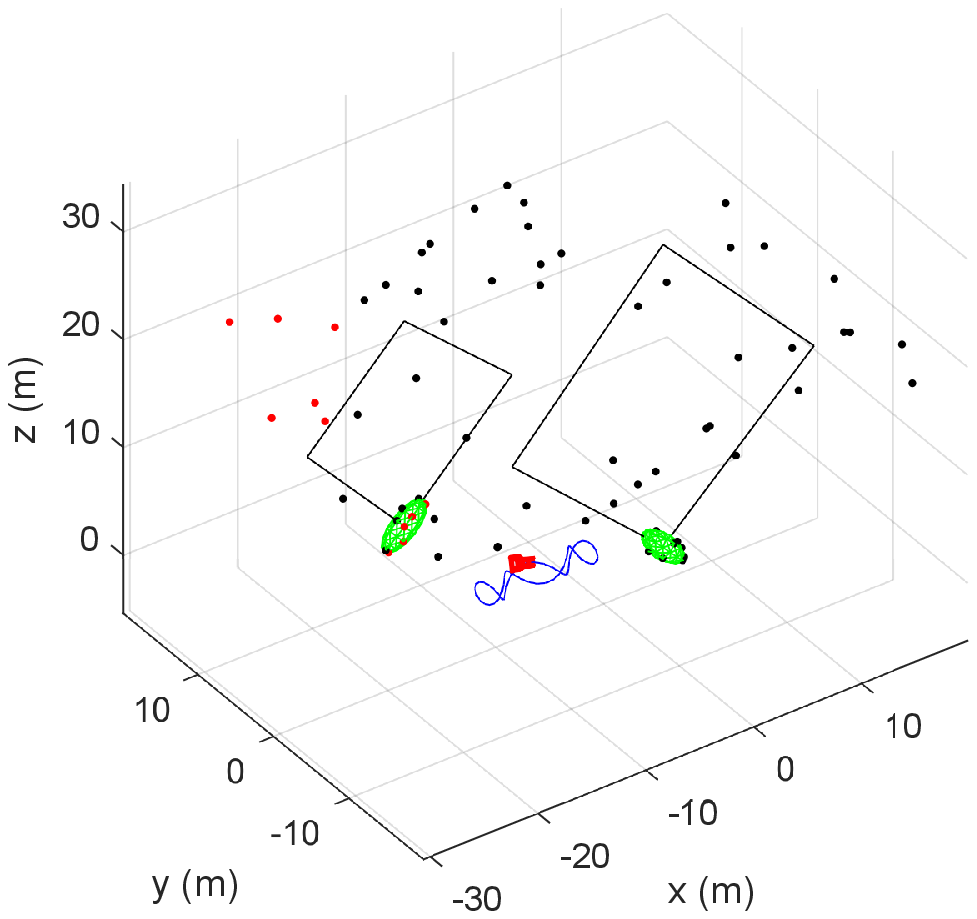}
		\caption{\small Experiment `D'}
		\label{fig:ExpDEnv}
	\end{subfigure}
	\begin{subfigure}[t]{0.19\textwidth}
		\centering
		\includegraphics[height=1.4in,trim=5cm 9cm 5cm 9cm ,clip]{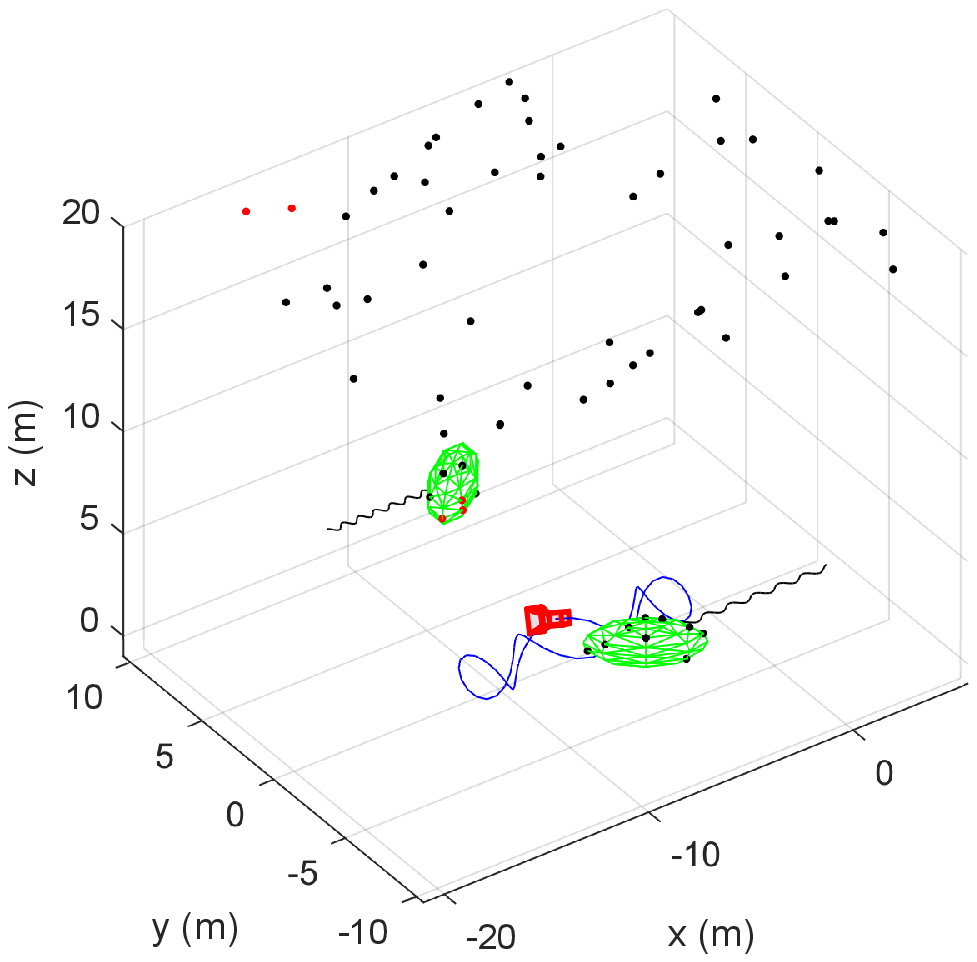}
		\caption{\small Experiment `E'}
		\label{fig:ExpEEnv}
	\end{subfigure}
	\caption{Different simulated robot and object trajectories. Robot trajectories are shown in blue and object trajectories in black.}
	\label{fig:robotObjTraj}
\end{figure*}

\section{Experimental methodology}
\label{sec:Setup}
%Overview
Our SLAM system is evaluated in a number of experiments with a robot moving in dynamic environments (with moving rigid body objects). The goal is to validate the proposed methodology and evaluate the benefits of integrating them into the SLAM problem. To this end, we used several simulated datasets with ground truth, and also tested our algorithm on a real dataset.

%Recently proposed methods for object detection and segmentation allow segmenting objects at image-level in real-time~\cite{Detectron2018,He17iccv,Redmon16cvpr}. \TODO{Make sure those are all real time}. 

A set of experiments were carried on simulated data obtained by emulating an advanced front-end that is capable of identifying objects, and associating detected landmarks with the different objects in the scene. In those simulations, two other types of measurements are also available; odometric measurements and point observations. Simulated data feature different robot and object trajectories of different sizes as seen in Figure \ref{fig:robotObjTraj}. A second, more realistic experiment, was performed on the Virtual KITTI dataset~\cite{Gaidon16cvpr}.
Furthermore, real data was acquired using a Turtlebot, equipped with an RGB-D camera, odometer and IMU, which moved in an environment where two other Turtlebots are moving to emulate dynamic objects. The setup was placed in a space with a VICON motion tracking system to provide the ground truth data.

\subsection{Simulated datasets}
\subsubsection{Experiment `A'}
\label{expA}
%% 168 experiments -- Experiment A
In order to test the effect of integrating
the objects' constant motion constraints, we designed an
experiment where a single object is being tracked by the robot,
and no static points are observed. Both robot and object are
following a circular trajectory, the object having a constant
motion. This experiment was repeated for $12$ different increasing lengths of the object and robot trajectories, yielding different number of ternary factors as defined in \eqref{eq:constantMotionGeneral} ranging from $34$ to $6715$. We will refer to this set of experiments as ``Experiment `A''' in the remainder of this document.

\subsubsection{Experiments `B' $\rightarrow$ `E'}
\label{expBE}
In order to evaluate the constant motion assumption, $4$ more experiments with multiple moving objects were generated using a simulated
environment. In the first two experiments of this set (we will refer to these as ``Experiments `B' and `C' in the remainder of this document), the objects are constrained to follow a
constant motion trajectory. In the experiment `B' the objects are following an elliptical trajectory as seen in Figure \ref{fig:ExpBEnv}, while in experiment `C', the objects are only translating in 3D as seen in Figure \ref{fig:ExpCEnv}. 

The constant motion assumption is violated in the following two experiments, in order to show the ability of our method to deal with non-constant motion. These experiments are referred to as``Experiments `D' and `E' '', and they feature objects following a rectangular trajectory in experiment D as seen in Figure \ref{fig:ExpDEnv} and objects following a sine wave trajectory in experiment E. We will refer to this set of experiments as ``Experiment `D' $\rightarrow$ `E'''. 
%\TODO{In the suplemetary material we add two more experiments with four objects: the experiment B-2 is similar to the experiment B with double the number of moving objects and the experiment D-2 is similar to the experiment D with double the number of moving objects.}
These experiments were then repeated with static points in the background to show the effect of integrating object tracking into our SLAM framework,
and demonstrate the improved accuracy of the estimation even when static landmarks are available.
% \TODO {this has to go in here} %This is made available in the submission supplementary material.

%Virtual KITTI dataset
\subsection{Virtual KITTI dataset}
\begin{figure}[h]
	\centering
	\includegraphics[width=0.9\linewidth,trim= 6cm 11cm 6cm 11cm,clip]{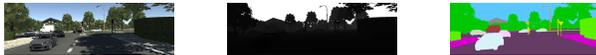}
	\caption{Examples of RGB, depth and segmentation images from the virtual Kitti dataset.}
	\label{fig:vKitti}
\end{figure}

\begin{figure}[h]
	\centering
	\includegraphics[width=0.9\linewidth]{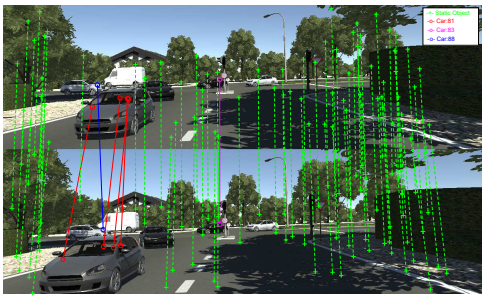}
	\caption{Feature extraction and tracking applied to the virtual kitti dataset. Static points are shown in green, and a different colour is used for points attached on each unique object.}
	\label{fig:vKittiFrontEnd}
\end{figure}

Virtual KITTI~\cite{Gaidon16cvpr} is a photo-realistic synthetic dataset designed to evaluate computer vision scene understanding algorithms. It contains $50$ high-resolution monocular videos generated from five different virtual worlds in urban settings under different imaging and weather conditions. These worlds were created using the Unity game engine and a
novel real-to-virtual cloning method. The photo-realistic synthetic videos are fully annotated at the pixel level with object labels as shown in Figure \ref{fig:vKitti}. The depth map of each image is also available. This makes it a perfect dataset to test and evaluate the proposed technique on realistic scenarios.
%x
%A feature matching and tracking algorithm was run to generate a set of 3D point measurements. This was complemented by the object segmentation, 
The front-end detects features in each image and obtain the 3D position relative to the camera of each point, with the aid of the corresponding depth images. The pixel-level object tracking is used to determine if these points are located on moving or static objects, and the known camera poses between frames are used to project the points to the next image in the sequence. In this manner, static and moving points are tracked between images to provide data associations between landmarks and objects as shown in Figure~\ref{fig:vKittiFrontEnd}. It is possible to project points attached to moving objects to other images as the pose of all moving objects is provided by the dataset for each image. As the 3D position is tracked for all the points, along with the camera pose, the relative position of all points can be obtained for every images for which the point remains in the camera's field of view. Upon completion of the point tracking through the chosen image set, the camera pose and relative point position are used to generate ground truth and measurement files. The noise levels added to the ground truth data in order to generate noisy measurements are as follows: $\Sigma_v = \diag[0.4m, 0.4m, 0.4m, 6^\circ, 6^\circ, 6^\circ]^2$ , $\Sigma_w = \diag[0.4m, 0.4m, 0.4m]^2$ , $\Sigma_q = \diag [0.05m, 0.05m, 0.05m]^2$. 
Our SLAM implementation can directly use these files, allowing it to be validated against the ground truth provided by the dataset. This experiment will be referred as ``vKITTI'' the remainder of this paper.

%Real Dataset
\subsection{Real dataset}
\label{realData}
The real data, was acquired using a Turtlebot equipped with an RGB-D Kinect sensor. Two other Turtlebots were covered with boxes, on which distinct colour coded landmarks were printed in order to facilitate landmark extraction, tracking and landmark/object association. Feature points landmarks on the walls and floor of the experiment space are also included in the estimation. The whole experiment is carried out in a space that is monitored by a motion capture and tracking system VICON. Three VICON markers were installed on each of the two objects and the camera robot to provide ground truth data for error analysis. This experiment will be referred to as ``real experiment'' in the remainder of this paper. We model the noise levels in the different sensors as follows: odometry measurement noise $\Sigma_v = \diag[0.04 m, 0.04 m, 0.04 m, 2^{\circ}, 2^{\circ}, 2^{\circ}]^2$, point measurement noise $\Sigma_w = \diag[0.2 m, 0.2 m, 0.2 m]^2$, and point-SE3 motion edge noise $\Sigma_q = \diag[0.05 m, 0.05 m, 0.05 m]^2$.

%\begin{figure}[h]
%	\centering
%	\includegraphics[width=0.9\linewidth,trim=0mm 30mm 0mm 30mm,clip]{./figures/realDataSetup.jpg}
%	\caption{A snapshot of the real data collection; two ``objects'' turtlebots were covered with boxes whose sides are painted in black and on which patches of distinct colour coded landmarks were added to ease the landmark extraction and labelling, and are being tracked by a turtlebot that acts as the ``camera''. All 3 turtlebots had VICON markers installed on them, to act as ground truth data.}
%	\label{fig:realDataSetup}
%\end{figure}

\subsection{Implementation details}
This work was implemented into a Matlab framework which is able to integrate not only simple point measurements but also additional available
information about the environment into a single SLAM framework. The object oriented design is thought to accommodate different types of information about the environment as long as there
is a front-end that can provide this information and a function that can model it. 
The framework consists of: 
\begin{inparaenum}
        \item a simulation component that can reproduce several dynamic environments;
	\item a front-end that generates the data for the SLAM problem by tracking features, objects and providing point associations using simulated or real data inputs; %component different simulation front-ends that feature realistic visual sensors, motion models, trajectories and environments,
	\item a back-end component that includes different non-linear solvers for batch and incremental processing.
\end{inparaenum}  
The estimation is implemented as a solution to an NLS problem as presented in section \ref{sec:Problem} and solved using Levenberg-Marquardt method.
The proposed technique can be easily integrated into any of the existing SLAM back-ends ~\cite{Kummerle11icra,Kaess11ijrr,Ila17ijrr}. The code for SLAM with dynamic objects will be made publicly available upon acceptance.

\section{Experimental results}
\label{sec:Experiments}
This section evaluates the proposed technique on the applications described in section \ref{sec:Setup}. We are focused on analysing the accuracy and consistency of the proposed estimation solution and comparing it to the classical SLAM formulation which does not integrate any additional information about the motion of the 3D points in the environment.

The accuracy of the solution of the SLAM problem is evaluated by comparing the absolute trajectory translational error (ATE), the absolute trajectory rotational error (ARE), the absolute structure error (ASE), the all to all relative trajectory translational error (allRTE), the all-to-all relative trajectory rotational error  (allRRE), and the all-to-all relative structure error (allRSE) calculated as in ~\cite{Ila17ijrr}. 

%%%%%%%%%%%%%%%%%%%%%%%%%%%%%%%%%%%%%%%
\subsection{Analysis of the simulated datasets} \label{subsec:simulatedDataAnalysis}

\begin{figure}
    	\begin{subfigure}[b]{0.24\textwidth}
    		\centering
    		\includegraphics[width=\textwidth,trim=0cm 0cm 0cm 0cm,clip]{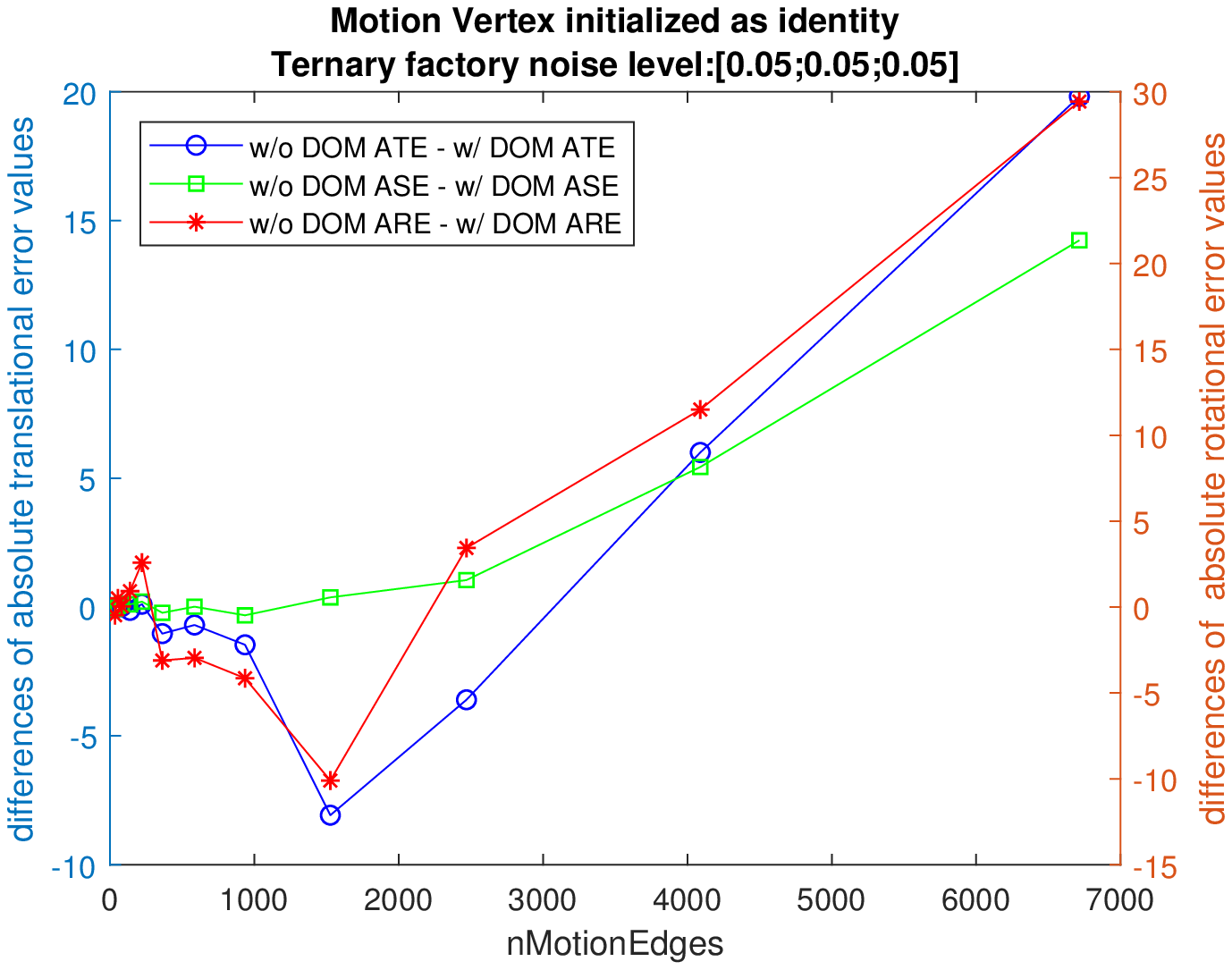}
    		\caption{{\scriptsize Absolute error differences.}}    
    		%\label{fig:Abs_Id}
    	\end{subfigure}
    	\begin{subfigure}[b]{0.24\textwidth}  
    		\centering 
    		\includegraphics[width=\textwidth,trim=0cm 0cm 0cm 0cm,clip]{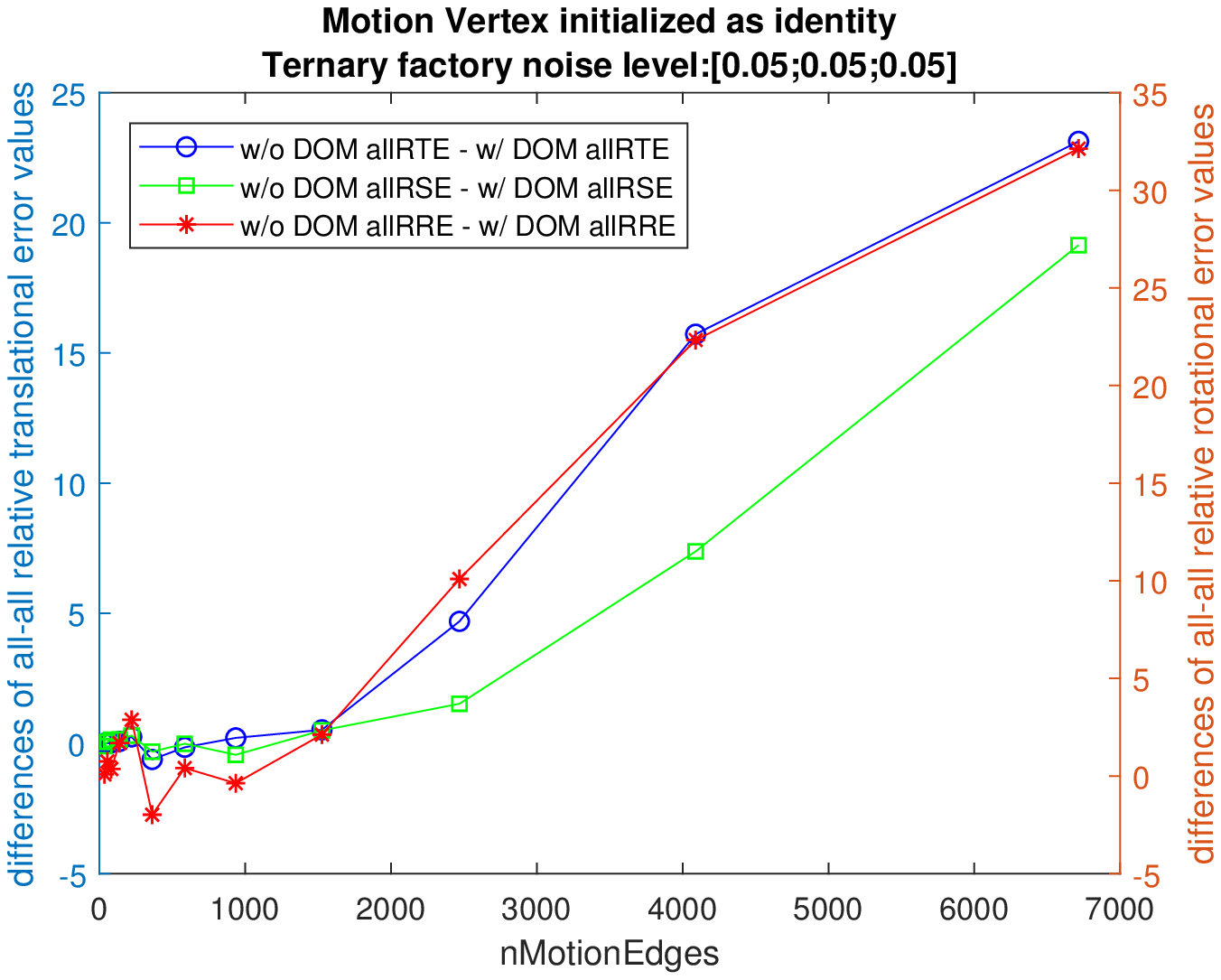}
    		\caption{{\scriptsize Relative error differences.}}    
    		%\label{fig:Rel_Id}
    	\end{subfigure}
%    	\begin{subfigure}[b]{0.245\textwidth}   
%    		\centering 
%    		\includegraphics[width=\textwidth,trim=4cm 9cm 4cm 9cm,clip]{./figures/Absolute_Tr}
%    		\caption{{\scriptsize Absolute errors difference when motion vertex is initialised as translation}}    
%    		\label{fig:Abs_Tr}
%    	\end{subfigure}
%    	\begin{subfigure}[b]{0.245\textwidth}   
%    		\centering 
%    		\includegraphics[width=\textwidth,trim=4cm 9cm 4cm 9cm,clip]{./figures/Relative_Tr}
%    		\caption{{\scriptsize Relative errors difference when motion vertex is initialised as translation}}    
%    		\label{fig:Rel_Tr}
%    	\end{subfigure}
    	\caption{\small Relative and absolute error differences computed as error metric of the SLAM solution without constant motion minus the same error metric of the SLAM solution integrating constant motion. The `x' axis represents the increasing number of ternary factors.}
    	\label{fig:AbsRelExpA}
    \end{figure}

\begin{figure*}[t]
\centering
\begin{tabular}{ccc c}
	\includegraphics[width=0.22\linewidth,trim={0cm 0cm 0cm 0cm},clip]{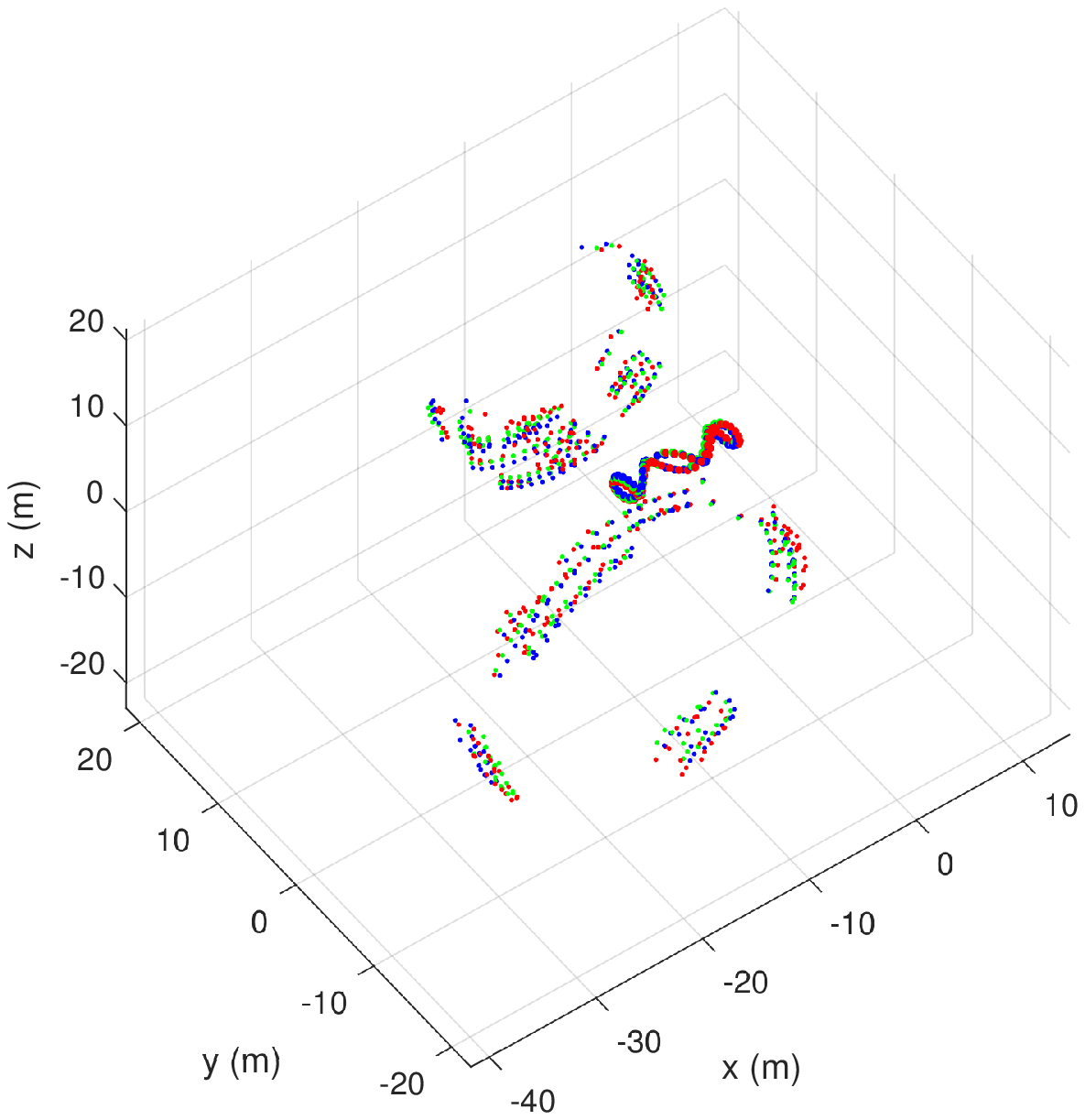} &
	\includegraphics[width=0.22\linewidth,trim={0cm 0cm 0cm 0cm},clip]{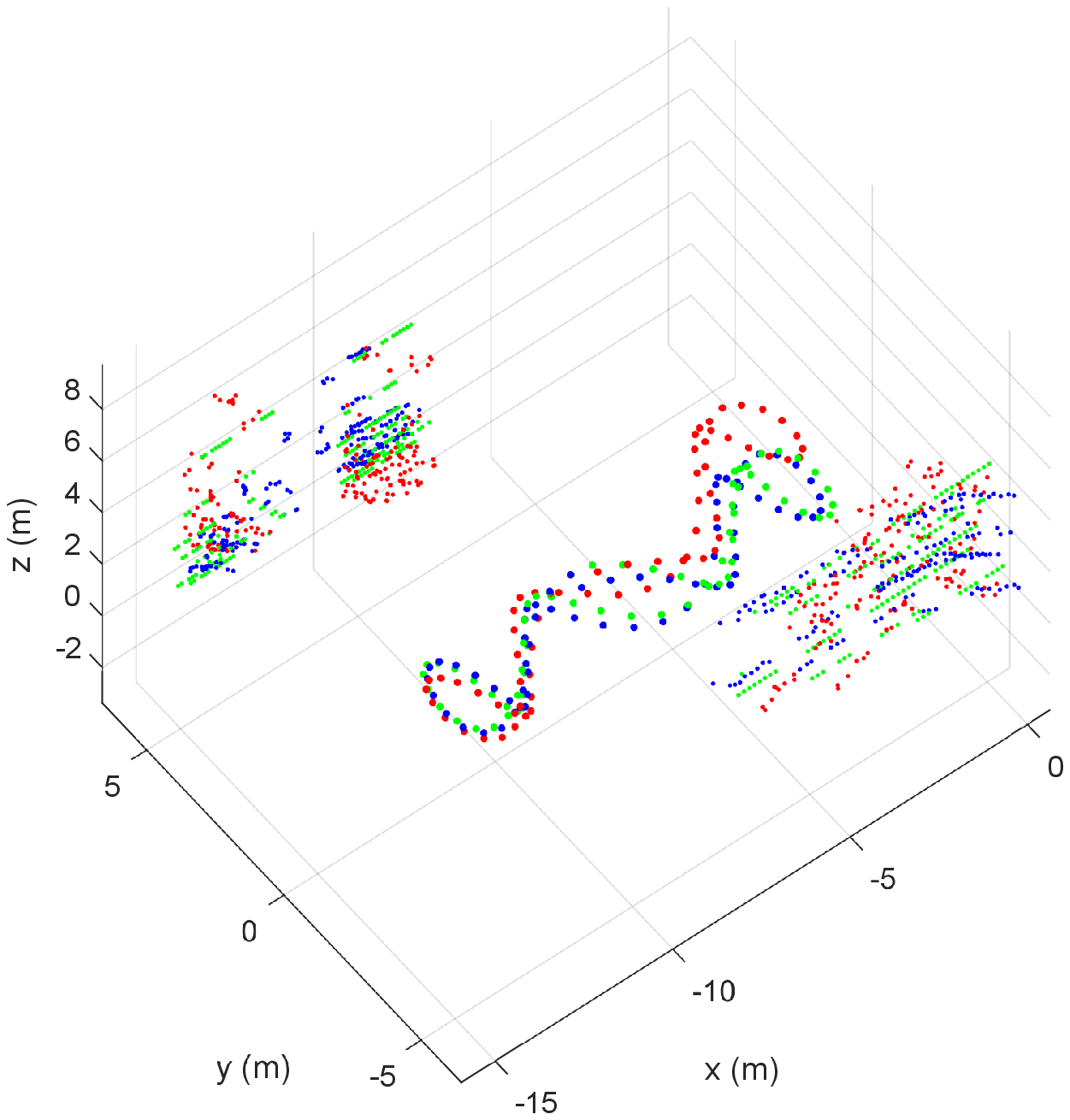}&
	\includegraphics[width=0.22\linewidth,trim={0cm 0cm 0cm 0cm},clip]{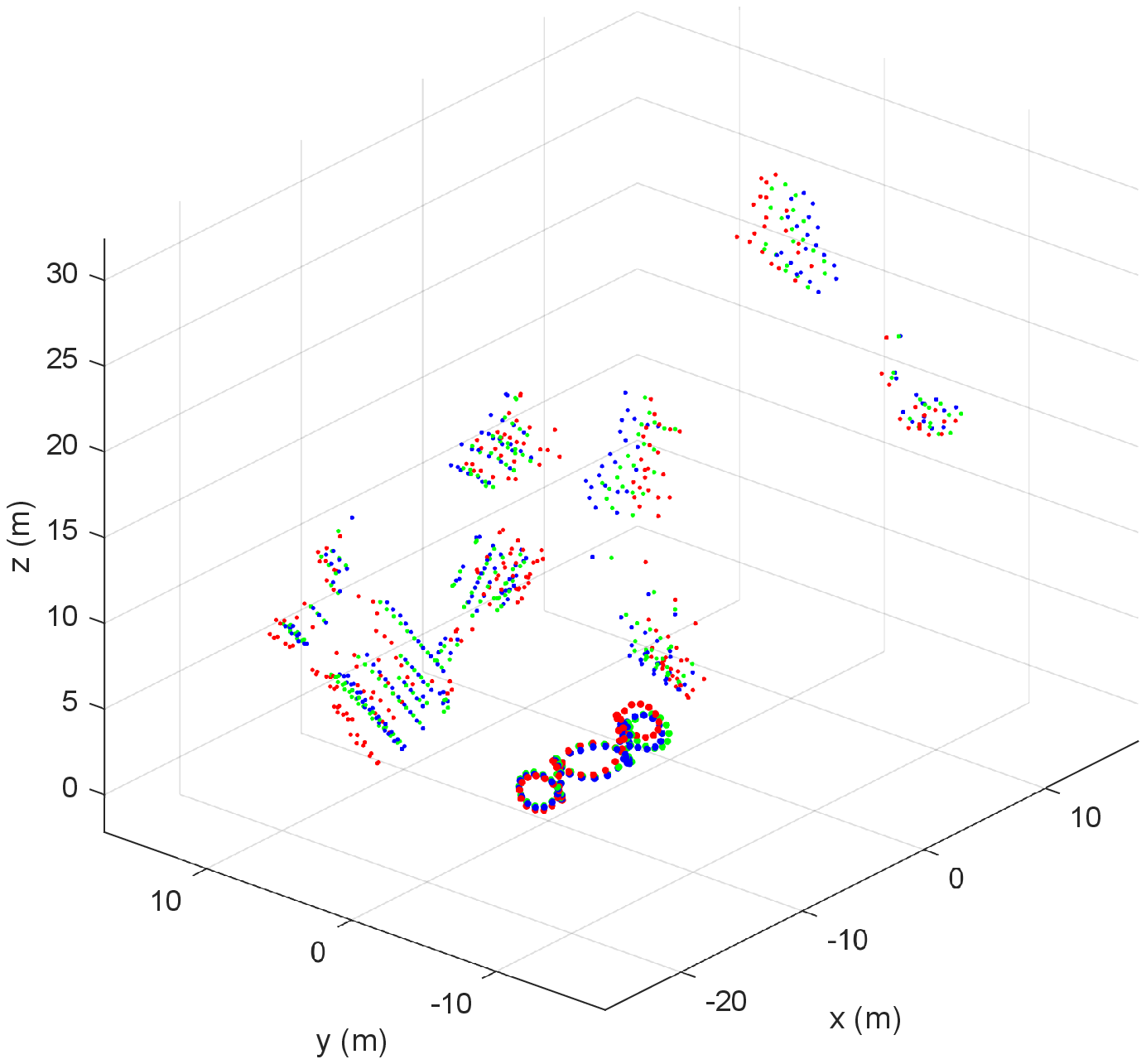}&
	\includegraphics[width=0.22\linewidth,trim={0cm 0cm 0cm 0cm},clip]{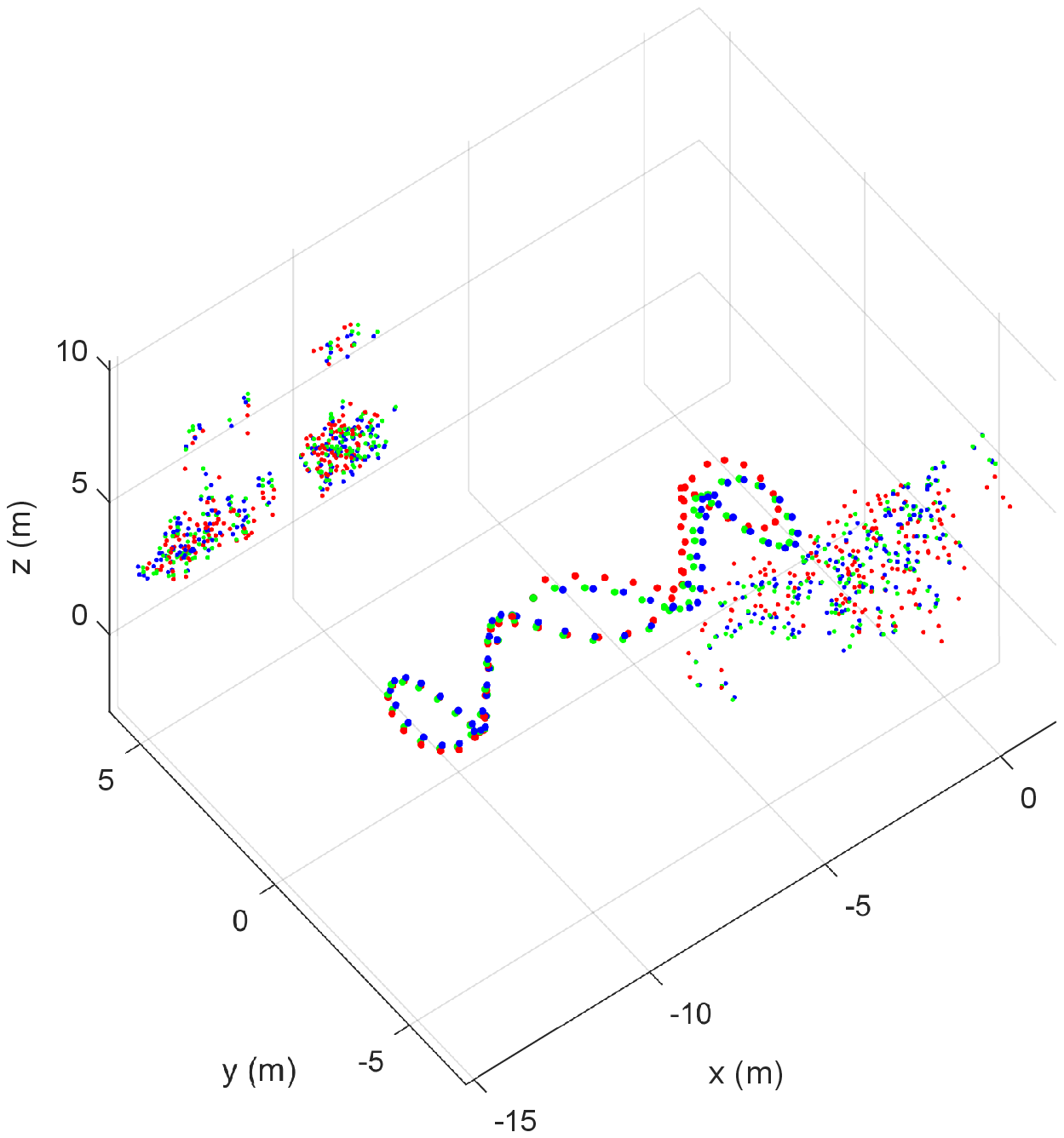}\\
	(a) Experiment `B' & (b) Experiment `C' & (c) Experiment `D' & (d) Experiment `E'
\end{tabular}
\caption{Solution of the SLAM with dynamic objects. Large dots represent robot trajectory and small dots represent moving points trajectories. Green - ground truth, Red - the SLAM solution without incorporating object motion information and Blue - the SLAM solution with constant motion. (best viewed in colour)}
\label{fig:expBE}
\end{figure*}

The tests show that the proposed method helps preserving the consistency of the map and improves the estimation quality significantly. This can be seen in Figure \ref{fig:AbsRelExpA} to Figure \ref{fig:expBE} and in Table \ref{tab:ExpA} to Table \ref{tab:DEResults}.

 %%%%%%%%%%%%%%%%%%%%%%%%%%%%%%%%%%%%%%%%%%%%%   
\begin{table}[h]
	\begin{center}
		\setlength\aboverulesep{0.1pt} \setlength\belowrulesep{0.1pt}
		\setlength\extrarowheight{1.8pt}
		\begin{tabularx}{8cm}{
				|n!{\vrule width\lightrulewidth}
				s!{\vrule width\lightrulewidth}
				s!{\vrule width\lightrulewidth}
				s!{\vrule width\lightrulewidth}
				s!{\vrule width\lightrulewidth}
				s!{\vrule width\lightrulewidth}
				s!{\vrule width\lightrulewidth}
				s!{\vrule width\lightrulewidth}s}
			\hline
			& \multicolumn{3}{c|} {\textbf{Experiment A}} \\   \hline
			\textbf{Average errors} & \textbf{w/o DOM} & \textbf{w/ DOM} & \textbf{\%}\\ \hline
			\textbf{ATE (m)} & 13.834 & \textbf{13.58} & \textcolor{blue}{1.82}\\ \hline
			\textbf{ARE ($^\circ$)} & 16.21 & \textbf{13.75} & \textcolor{blue}{15.1} \\ \hline
			\textbf{ASE (m)} & 3.93 & \textbf{1.426} & \textcolor{blue}{63.7} \\ \hline
			\textbf{allRTE (m)} & 8.249 & \textbf{3.002} & \textcolor{blue}{63.6} \\ \hline
			\textbf{allRRE ($^\circ$)} & 14.103 & \textbf{6.859} & \textcolor{blue}{51.3} \\ \hline
			\textbf{allRSE (m)} & 5.408 & \textbf{2.015} & \textcolor{blue}{62.7} \\ \hline
		\end{tabularx}
	\end{center}
	\caption{Average error values for a total of 13 experiments with the motion vertex initialised as identity. `w/ DOM' denotes the proposed estimation technique with dynamic object motion integrated, while `w/o DOM' means that no object motion information was used in the estimation. A positive \% shows improvement of `w/ DOM'.}
	\label{tab:ExpA}
\end{table}
    
  \subsubsection{Experiment `A'}
The results in Table \ref{tab:ExpA}  and Figure \ref{fig:AbsRelExpA} show improvements of the estimation accuracy when the objects' motion is integrated into the estimation. Several increasing in size SLAM problems were evaluated and the relative and absolute error differences of the pose of the robot and the location of the 3D points are reported. Positive values indicate better performance of the SLAM problem which integrate constant motion information. The general trend of the results shows that adding the motion information has a positive effect on the estimation in most of the cases. However, there are few cases where adding this constant motion vertex and the ternary factors does not benefit the solution. This is probably due to added non-linearities by the $\se$ type vertex which affects the convergence. Note that this experiment is showing an extreme case, where the environment has only one moving object and no static points.

\begin{table}[t]
	\begin{center}
		\setlength\aboverulesep{0.1pt} \setlength\belowrulesep{0.1pt}
		\setlength\extrarowheight{1.8pt}
		\begin{tabularx}{8cm}{
				|n!{\vrule width\lightrulewidth}
				s!{\vrule width\lightrulewidth}
				s!{\vrule width\lightrulewidth}
				s!{\vrule width\lightrulewidth}
				s!{\vrule width\lightrulewidth}
				s!{\vrule width\lightrulewidth}
				s!{\vrule width\lightrulewidth}
				s!{\vrule width\lightrulewidth}s}
			\hline
			& \multicolumn{3}{c|} {\textbf{Exp.B}}  & \multicolumn{3}{c|} {\textbf{Exp.C}} \\  \hline
			\multicolumn{7}{|c|} {\textbf{with no static points}}\\  \hline
			\textbf{Error} & \textbf{w/o DOM} & \textbf{w/ DOM} & \textbf{\%} & \textbf{w/o DOM} & \textbf{w/ DOM}  & \textbf{\%}\\ \hline
			\scriptsize \textbf{ATE (m)} & 0.342 & \textbf{0.203} & \textcolor{blue}{40.6} & 0.453 & \textbf{0.351} & \textcolor{blue}{22.5}\\ \hline
			\scriptsize \textbf{ARE ($^\circ$)} & 6.211 & \textbf{3.751} & \textcolor{blue}{39.6} & 6.371 & \textbf{4.644}  & \textcolor{blue}{27.1} \\ \hline
			\scriptsize \textbf{ASE (m)} & 0.733 & \textbf{0.498} & \textcolor{blue}{32.1} & 0.567 & \textbf{0.319} & \textcolor{blue}{43.7} \\ \hline
			\tiny \textbf{allRTE (m)} & 0.213 & \textbf{0.171} & \textcolor{blue}{19.7} & 0.393 & \textbf{0.236} & \textcolor{blue}{39.9}\\ \hline
			\tiny \textbf{allRRE ($^\circ$)} & 5.183 & \textbf{3.665} & \textcolor{blue}{29.3} & 5.410 & \textbf{5.012} & \textcolor{blue}{7.3} \\ \hline
			\tiny \textbf{allRSE (m)} & 1.049 & \textbf{0.707} & \textcolor{blue}{32.6} & 0.797 & \textbf{0.439} & \textcolor{blue}{44.9} \\ \hline
			\multicolumn{7}{|c|} {\textbf{with static points}}\\ \hline
			\scriptsize \textbf{ATE (m)} & 0.285 & \textbf{0.278} & \textcolor{blue}{2.45} & 0.578 & \textbf{0.534} & \textcolor{blue}{7.61}\\ \hline
			\scriptsize \textbf{ARE ($^\circ$)} & 2.950 & \textbf{1.229} & \textcolor{blue}{58.3} & \textbf{7.381} & 7.508  & \textcolor{red}{-1.7} \\ \hline
			\scriptsize \textbf{ASE (m)} & 1.031 & \textbf{0.327} & \textcolor{blue}{68.3} & 0.993 & \textbf{0.478} & \textcolor{blue}{51.8} \\ \hline
			\tiny \textbf{allRTE (m)} & 0.359 & \textbf{0.193} & \textcolor{blue}{46.2} & 0.212 & \textbf{0.175} & \textcolor{blue}{17.5}\\ \hline
			\tiny \textbf{allRRE ($^\circ$)} & 3.415 & \textbf{1.423} & \textcolor{blue}{58.3} & 4.867 & \textbf{3.377} & \textcolor{blue}{30.6} \\ \hline
			\tiny \textbf{allRSE (m)} & 1.422 & \textbf{0.438} & \textcolor{blue}{69.2} & 1.419 & \textbf{0.715} & \textcolor{blue}{49.6} \\ \hline
		\end{tabularx}
	\end{center}
	\caption{Error values for experiments `B' and `C' explained in section \ref{expBE}.  `w/ DOM' denotes the proposed estimation technique with dynamic object motion, while `w/o DOM' means that no object motion information was used in the estimation. A positive \% shows improvement of `w/ DOM'.}
	\label{tab:BCResults}
\end{table}

\begin{table}[t]
	\begin{center}
		\setlength\aboverulesep{0.1pt} \setlength\belowrulesep{0.1pt}
		\setlength\extrarowheight{1.8pt}
		\begin{tabularx}{8cm}{
				|n!{\vrule width\lightrulewidth}
				s!{\vrule width\lightrulewidth}
				s!{\vrule width\lightrulewidth}
				s!{\vrule width\lightrulewidth}
				s!{\vrule width\lightrulewidth}
				s!{\vrule width\lightrulewidth}
				s!{\vrule width\lightrulewidth}
				s!{\vrule width\lightrulewidth}s}
			\hline
			& \multicolumn{3}{c|} {\textbf{Exp.D}}  & \multicolumn{3}{c|} {\textbf{Exp.E}} \\  \hline
			\multicolumn{7}{|c|} {\textbf{with no static points}}\\ \hline
			\textbf{Error} & \textbf{w/o DOM} & \textbf{w/ DOM} & \textbf{\%} & \textbf{w/o DOM} & \textbf{w/ DOM}  & \textbf{\%}\\ \hline
			\scriptsize \textbf{ATE (m)} & 0.457 & \textbf{0.207} & \textcolor{blue}{54.7} & 0.549 & \textbf{0.314} & \textcolor{blue}{42.8}\\ \hline
			\scriptsize \textbf{ARE ($^\circ$)} & 6.439 & \textbf{4.248} & \textcolor{blue}{34} & 6.085 & \textbf{2.249}  & \textcolor{blue}{63} \\ \hline
			\scriptsize \textbf{ASE (m)} & 1.127 & \textbf{0.595} & \textcolor{blue}{47.2} & 0.546 & \textbf{0.142} & \textcolor{blue}{73.9} \\ \hline
			\tiny \textbf{allRTE (m)} & \textbf{0.258} & 0.314 & \textcolor{red}{-21.7} & 0.234 & \textbf{0.185} & \textcolor{blue}{20.9}\\ \hline
			\tiny \textbf{allRRE ($^\circ$)} & 5.719 & \textbf{4.180} & \textcolor{blue}{26.9} & 5.655 & \textbf{2.674} & \textcolor{blue}{52.7} \\ \hline
			\tiny \textbf{allRSE (m)} & 1.505 & \textbf{0.811} & \textcolor{blue}{46.1} & 0.73 & \textbf{0.202} & \textcolor{blue}{72.3} \\ \hline
			\multicolumn{7}{|c|} {\textbf{with static points}}\\ \hline
			\scriptsize \textbf{ATE (m)} & 0.322 & \textbf{0.123} & \textcolor{blue}{61.8} & 0.308 & \textbf{0.143} & \textcolor{blue}{53.6}\\ \hline
			\scriptsize \textbf{ARE ($^\circ$)} & 4.910 & \textbf{2.252} & \textcolor{blue}{54.1} & 4.876 & \textbf{1.425}  & \textcolor{blue}{70.7} \\ \hline
			\scriptsize \textbf{ASE (m)} & 1.255 & \textbf{0.422} & \textcolor{blue}{66.4} & 0.733 & \textbf{0.240} & \textcolor{blue}{67.3} \\ \hline
			\tiny \textbf{allRTE (m)} & 0.325 & \textbf{0.208} & \textcolor{blue}{36} & 0.320 & \textbf{0.178} & \textcolor{blue}{44.4}\\ \hline
			\tiny \textbf{allRRE ($^\circ$)} & 4.240 & \textbf{1.998} & \textcolor{blue}{52.9} & 4.854 & \textbf{1.946} & \textcolor{blue}{59.9} \\ \hline
			\tiny \textbf{allRSE (m)} & 1.719 & \textbf{0.576} & \textcolor{blue}{66.5} & 1.017 & \textbf{0.348} & \textcolor{blue}{65.8} \\ \hline
		\end{tabularx}
	\end{center}
	\caption{Error values for experiments `D' and `E' explained in section \ref{expBE}.  `w/ DOM' denotes the proposed estimation technique with dynamic object motion, while `w/o DOM' means that no object motion information was used in the estimation. A positive \% shows improvement of `w/ DOM'.}
	\label{tab:DEResults}
\end{table}

%	\caption{Experiment `B' results. Large dots represent robot positions and small dots point locations. Green - ground truth, red - the SLAM solution without incorporating object motion information and blue represents the SLAM solution with motion integrated.(best viewed in colour)}
%	\label{fig:expBResults}
%\end{figure}
%
%	\caption{Experiment `D' results. Large dots represent robot positions and small dots point locations. Green - ground truth, red - the SLAM solution without incorporating object motion information and blue represents the SLAM solution with motion integrated.(best viewed in colour)}
%	\label{fig:expCResults}
%\end{figure}
%	\caption{Experiment `D' results. Large dots represent robot positions and small dots point locations. Green - ground truth, red - the SLAM solution without incorporating object motion information and blue represents the SLAM solution with motion integrated.(best viewed in colour)}
%	\label{fig:expDResults}
%\end{figure}
%
%\begin{figure}[h]`A'
%	\centering
%
%	\caption{Experiment `E' results. Large dots represent robot positions and small dots point locations. Green - ground truth, red - the SLAM solution without incorporating object motion information and blue represents the SLAM solution with motion integrated.(best viewed in colour)}

\subsubsection{Experiments `B' $\rightarrow$ `E'}

The upper part of the Tables  \ref{tab:BCResults} and \ref{tab:DEResults} show the accuracy results for the Experiments  `B' $\rightarrow$ `E' with no static points.  The values indicate that adding information about the motion of the objects in the estimation process significantly improves the estimation quality and reduces the trajectory and map errors in both absolute and relative metrics. This can be seen in the positive $\%$ values in the tables. Surprisingly, the improvement persists in Experiments `D' and `E' where the constant motion assumption is slightly violated. The explanation for that is that the estimator compromises in estimating the constant pose transformation and satisfying the point measurements in both cases. Same noise levels were used for the ternary factors, the point measurements and the odometry across the four experiments. Qualitative evaluation of the same experiments is shown in Figure \ref{fig:expBE}. Note that the ground truth (in green) and the SLAM accounting for constant motion (in blue) have similar values, while without accounting for the constant motion (in red) diverges from the ground truth. 

The lower part of the Tables \ref{tab:BCResults} and \ref{tab:DEResults} show the accuracy results for the Experiments  `B' $\rightarrow$ `E' where static points were also integrated into the system. We note here overall increase in accuracy due to the know effect of the static points data association on the SLAM solution.

%Results from this set of experiments show that the proposed method significantly improves the estimation quality and reduces the trajectory and map errors in both absolute and relative.  This can be seen in Figure \ref{fig:expBResults} to Figure \ref{fig:expBE}.
%
%The proposed method reduces the drift that accumulates on the robot poses, while improving the structure estimation. This is also visible in the error values in Table \ref{tab:BCResults} and Table \ref{tab:DEResults}.
%
%It is consistently visible across all the experiment that the improvements are mostly happening in the points error, which are mostly affected by the new vertices and edges added to the graph, and that act directly on them. This can be in every table in this paper.
%
%
%\todos{I am not sure I perfectly understand the fact that when the motion is not perfectly constant in SE3, the percentage improve is much higher when modelled as SE3!! The only explanation I can think of, is the following; when the motion is perfectly an SE3 motion, and the degree of confidence I assign to the 2points-constant motion vertex is std = diag([0.05,0.05,0.5]), it's hard for the system to find a perfect H to fit the motion of the points, due to the noise in point-measurements and robot poses. However, when the motion, is close to SE3, but not perfectly an SE3 motion, the system has some degree of freedom to play within, and is easier to find a transformation H, that fits the noisy points and at the same time the 2points-SE3 motion edges introduced. }

%%%%%%%%%%%%%%%%%%%%%%%%%%%%%%%%%%%%%%%%%%%%%   

\subsubsection{Experiment ``vKITTI''}
An experiment was run over a total of $295$ vertices, including robot poses, static and dynamic landmark positions and constant motion vertices resulting in a total $508$ edges, of which $216$ are object motion related ternary factors. The bottom part of the Table~\ref{tab:realDataResults} shows the accuracy of the SLAM solution obtained with and without constant motion information. As expected, the solution of the estimation improves when accounting for moving objects in the scene. %Note that in this experiment is 

%Figure \ref{fig:vKittiResults} shows the final estimate of the vKITTI dataset, with and without integration of dynamic object motion into the estimation. 
%
%\begin{figure}[h]
%	\centering
%	%\includegraphics[width=0.8\linewidth,trim={3.5cm 7.8cm 3.5cm 9cm},clip]{./figures/vKitti_Solution.pdf}
%	\caption{Top view of the final vKITTI dataset estimate with no motion integrated (left) vs with objects' motion integrated  (right). }
%	\label{fig:vKittiResults}
%\end{figure}

%%%%%%%%%%%%%%%%%%%%%%%%%%%%%%%%%%%%%%%%%%
\begin{table}[h]
	\begin{center}
		\setlength\aboverulesep{0.1pt} \setlength\belowrulesep{0.1pt}
		\setlength\extrarowheight{1.8pt}
		\begin{tabularx}{8cm}{
				|n!{\vrule width\lightrulewidth}
				s!{\vrule width\lightrulewidth}
				s!{\vrule width\lightrulewidth}
				s!{\vrule width\lightrulewidth}
				s!{\vrule width\lightrulewidth}
				s!{\vrule width\lightrulewidth}
				s!{\vrule width\lightrulewidth}
				s!{\vrule width\lightrulewidth}s}
			\hline
			& \multicolumn{3}{c|} {\textbf{Real data Exp.}} \\   \hline
			\textbf{Error} & \textbf{w/o DOM} & \textbf{w/ DOM} &  \textbf{\%}\\ \hline
			\textbf{ATE (m)} & 1.112 & \textbf{0.981} & \textcolor{blue}{11} \\ \hline
			\textbf{ARE ($^\circ$)} & 28.021 & \textbf{27.654} & \textcolor{blue}{1} \\ \hline
			\textbf{ASE (m)} & 1.195 & \textbf{0.963} & \textcolor{blue}{19} \\ \hline
			\textbf{allRTE (m)} & 0.025 & \textbf{0.023} & \textcolor{blue}{19} \\ \hline
			\textbf{allRRE ($^\circ$)} & 1.681 & \textbf{1.533} & \textcolor{blue}{8} \\ \hline
			\textbf{allRSE (m)x$10^{-3}$} & 0.042 & \textbf{0.035} & \textcolor{blue}{16\%} \\ \hline
			
			& \multicolumn{3}{c|} {\textbf{vKITTI Exp.}} \\   \hline
			\textbf{ATE (m)}           &    3.060 &  \textbf{1.439}&    \textcolor{blue}{52.9} \\ \hline
			\textbf{ARE ($^\circ$)} &    13.270 &  \textbf{ 3.970}& \textcolor{blue}{70.0} \\ \hline
			\textbf{ASE (m)}          &     1.864 &   \textbf{0.845}& \textcolor{blue}{54.7} \\ \hline
			\textbf{allRTE (m)}        &     3.055&    \textbf{ 0.874}& \textcolor{blue}{71.4} \\ \hline
			\textbf{allRRE ($^\circ$)} &     9.289&    \textbf{3.361}& \textcolor{blue}{63.8} \\ \hline
			\textbf{allRSE (m)x$10^{-3}$} &   2.736& \textbf{1.166} & \textcolor{blue}{ 57.4} \\ \hline
			
		\end{tabularx}
	\end{center}
	\caption{Result values for the real dataset and vKITTI experiments. ``w/ DOM'' denotes the proposed estimation technique with constrained object motion, while ``w/o'' means that no object motion information was used in the estimation. }
	\label{tab:realDataResults}
\end{table}

\subsection{Analysis of the real dataset}
\label{subsec:realDataAnalysis}
The real dataset is used to validate the algorithm on a real world scenario. However the problem size is small ($71$ vertices and $194$ edges, with the motion information integrated and only $69$ vertices and $165$ edges with no motion information added). % and does not showcase the amount of improvements brought to the estimation by incorporating objects' motion information. 
In general, the significant improvements are mostly in the 3D points estimation, and this is because the new constant motion vertex and the ternary factors act directly on the structure points. Importantly, the value of the noise level $\Sigma_q$ assigned to the motion edge impacts the solution significantly. Modelling using low noise values could be too risky, as in a real scenarios the object's motion is never constant, likewise high noise levels would result in an estimation that does not rely on the added motion information and would bring no benefit to the system.

%%%%%%%%%%%%%%%%%%%%%%%%% Conclusion
\section{Conclusion and future work}
\label{sec:Conclusion}
In this paper we proposed a new pose change representation for SLAM with dynamic objects and integrated it into a SLAM framework. The formulation has the advantage that no additional object pose or object geometry is required in order to account for the moving objects in SLAM. The framework works with simple SLAM front-ends capable of tracking feature points, detecting objects, and associating feature points to the objects in a sequence of images. Furthermore, we have analyzed the effect of integrating information about the objects' motion in the environment on the accuracy and consistency of the SLAM problem. In particular, we tested the constant motion assumption. Results show good improvements in the estimation quality and consistency of the result versus the same problem with no added object motion information.
%We showed that using an inertial frame coordinate representation of relative motion of the objects, eliminates the need of introducing object coordinates into the estimation process simplifying both the front-end, as well as the SLAM formulation. 
Although the examples presented in this paper use the prior information that the objects have a constant motion, the proposed formulation can easily be adapted to other types of motion and in the near future work we plan to examine different solutions.

Another important issue to be analyzed in the future is the computational complexity of SLAM with object dynamic objects. It is important to note that without further reductions, the problem can become intractable in large-scale environments with many moving objects. But at the same time state reduction can be easily implemented using a windowing approach that mantains only a small set of objects points in the state rather than the points observed along the entire trajectory. A principled way to do that is to analyze how much the old observation observations contribute to the solution of the SLAM problem~\cite{Kaess11ijrr,Polok13rss,Polok15acra}. In the future, we plan to restrict the optimization problem to relevant state space, and produce scalable solutions.

Finally we plan to tackle the problems of object segmentation and data association using state of the-art deep learning techniques \cite{Detectron2018,He17iccv,Redmon16cvpr} to achieve a full end-to-end SLAM system.

\section*{Acknowledgments}
This research was supported by the Australian Research Council through the ``Australian Centre of Excellence for Robotic Vision'' CE140100016. We would also like to thank Montiel Abello for his contribution with the initial framework implementation, and Yon Hon Ng, for his help collecting the real dataset.

%% Use plainnat to work nicely with natbib.

\bibliographystyle{plainnat}
\bibliography{biblio}

\end{document}